\documentclass[10pt]{article} 
\usepackage[preprint]{tmlr}

\usepackage{amsmath,amsfonts,bm}

\def\eqref#1{equation~\ref{#1}}

\def\1{\bm{1}}

\DeclareMathAlphabet{\mathsfit}{\encodingdefault}{\sfdefault}{m}{sl}
\SetMathAlphabet{\mathsfit}{bold}{\encodingdefault}{\sfdefault}{bx}{n}

\usepackage{hyperref}
\usepackage{url}
\usepackage{graphicx}
\usepackage{booktabs}
\usepackage{subcaption}
\usepackage{array}
\usepackage{orcidlink}

\usepackage{hyperref}
\usepackage{url}
\usepackage{tabularx}
\usepackage{graphicx}
\usepackage{tcolorbox} 
\usepackage{fontawesome5} 
\usepackage{booktabs}
\usepackage{multirow}
\usepackage{graphicx}
\usepackage{subcaption}
\usepackage{booktabs}
\usepackage{multirow}
\usepackage{colortbl}
\usepackage{xcolor}
\usepackage{arydshln}
\definecolor{lightblue}{RGB}{225,240,255}

\definecolor{dropred}{RGB}{200,30,30}

\usepackage{graphicx}
\usepackage{booktabs}
\usepackage[accsupp]{axessibility}
\usepackage{orcidlink}
\usepackage{hyperref}
\usepackage{url}
\usepackage{tabularx}
\usepackage{tcolorbox} 
\tcbuselibrary{skins, breakable} 
\usepackage{fontawesome5} 
\usepackage{multirow}
\usepackage{subcaption}
\usepackage{colortbl}
\usepackage{arydshln}
\usepackage{amsmath}

\definecolor{lightblue}{RGB}{225,240,255}
\definecolor{dropred}{RGB}{200,30,30}
\definecolor{impgreen}{RGB}{0,140,60}
\definecolor{ECCVBlue}{RGB}{0, 51, 102} 
\definecolor{SoftGray}{RGB}{245, 247, 250}
\definecolor{LinkColor}{RGB}{0, 85, 170}

\newcommand{\down}[1]{\textcolor{dropred}{\scriptsize($\downarrow$#1)}}

\renewcommand{\thesection}{\Alph{section}}

\usepackage{xcolor}
\usepackage{colortbl}
\definecolor{lightblue}{RGB}{225,240,255}
\definecolor{dropred}{RGB}{200,30,30}

\usepackage{xcolor}
\usepackage{colortbl}
\usepackage{amsmath}

\definecolor{lightblue}{RGB}{225,240,255}
\definecolor{dropred}{RGB}{200,30,30}
\definecolor{impgreen}{RGB}{0,140,60}

\usepackage{wrapfig}
\usepackage{hyperref}

\title{UHP Detection: LVLMs have their Unique Hallucination Pattern in the Consistency Space}

\author{\name Amir Mohammad Ezzati$^{\dagger}$
\addr Sharif University of Technology
\AND
\name Kiyan Rezaee$^{\dagger}$
\addr Michigan State University
\AND
\name Bardiya Kariminia
\addr Shahid Beheshti University
\AND
\name Mohamad Amin Yousefi
\addr University of Tehran
\AND
\name Asal Mohammadjafari Mamaqani
\addr Amirkabir University of Technology
\AND
\name Behrad Samimi
\addr Sharif University of Technology
\AND
\name Mohammad Hossein Rohban
\addr Sharif University of Technology
}

\def\month{MM}  
\def\year{YYYY} 
\def\openreview{\url{https://openreview.net/forum?id=XXXX}} 

\begin{document}

\maketitle

\begin{center}
\small
$^{\dagger}$Corresponding Authors:
\href{mailto:iamirezzati@gmail.com}{iamirezzati@gmail.com},
\href{mailto:rezaeeki@msu.edu}{rezaeeki@msu.edu}.
\end{center}

\begin{abstract}
Large vision--language models (LVLMs) demonstrate strong multimodal reasoning capabilities but remain prone to hallucination, where model predictions are not grounded in visual evidence. Existing black-box hallucination detection methods estimate uncertainty through a single consistency metric, implicitly assuming that model uncertainty can be adequately characterized by a single measure. However, hallucinations exhibit diverse manifestations of uncertainty across different behavioral probes, making a single measure insufficient to characterize their underlying behavior. We propose \emph{Unique Hallucination Pattern (UHP) Detection}, a fully black-box framework that models hallucination as a structured uncertainty pattern defined by two axes: perturbation modality (image vs.\ text) and logical polarity (a statement vs.\ its negation). Their intersection produces four complementary consistency groups that capture distinct manifestations of model uncertainty, from which both within-group and between-group features are extracted to train a lightweight classifier. Through comprehensive experiments on AMBER and PhD across three LVLMs, UHP Detection consistently outperforms prior black-box and white-box baselines, with improvements of up to $+18.72\%$ AUC-ROC and $+20.07\%$ AUC-PR over the strongest black-box methods. Extensive ablation studies demonstrate that each consistency group contributes complementary information and that their combination forms a structured hallucination pattern. Furthermore, cross-dataset evaluation shows that this learned pattern generalizes across benchmarks, indicating that hallucination behavior reflects a model-specific consistency pattern. \textbf{Code is publicly available at} \url{https://github.com/amirezzati/uhpdet}.
\end{abstract}

\section{Introduction}
\label{sec:intro}

Recent advances in Large Vision–Language Models (LVLMs)~\cite{bai2025qwen2, dai2023instructblip, zhu2023minigpt, chen2024internvl} have substantially improved multimodal reasoning, enabling image-conditioned question answering, caption generation, and visual inference. Despite their strong empirical performance, LVLMs remain prone to hallucination—producing responses that are factually incorrect or not grounded in the visual input. Such failures undermine reliability and raise particular concerns in high-stakes domains such as healthcare and autonomous systems, where plausible but incorrect outputs may influence downstream decisions~\cite{kim2025medical}.

Existing approaches to hallucination detection can be broadly classified into resource-dependent and resource-free methods~\cite{wang2023amber, liu2025phd, zhang2024vl, li2024reference, zhao2025visual}. Resource-dependent methods~\cite{wang2023amber, liu2025phd} use external resources—such as curated knowledge bases, vision models, or manually annotated datasets, to verify the consistency of model outputs with external evidence, but their effectiveness is limited by the availability and coverage of those resources. In contrast, resource-free methods~\cite{li2024reference, zhang2024vl, zhao2025visual} detect hallucinations without external resources, typically by exploiting signals already available within the model. A prominent line of work in this category estimates the model’s uncertainty in its outputs, under the assumption that uncertainty is informative about the presence of hallucinations~\cite{li2024reference, zhang2024vl, zhao2025visual}.

Within resource-free methods, uncertainty-based detection can be further divided into white-box and black-box techniques. White-box methods~\cite{zhao2025visual} analyze internal model signals—such as logits, parameters, or hidden states—to estimate uncertainty. These methods are limited to open-source models, where such internal information is accessible. Black-box methods~\cite{li2024reference, zhang2024vl}, however, derive uncertainty signals solely from observable model outputs, such as response consistency or answer variability. These techniques are applicable to both open- and closed-source models, making them more widely deployable in real-world settings.

\begin{wrapfigure}{r}{0.55\linewidth}
  \centering
  \includegraphics[width=\linewidth]{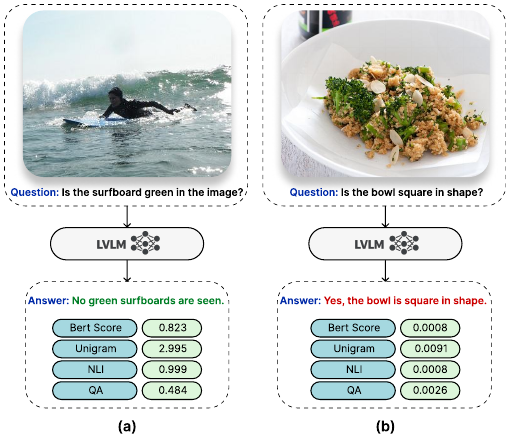}
  \caption{\textbf{Failure of inconsistency metrics}. (a) A correct response exhibits high inconsistency (uncertainty), while (b) a hallucinated response demonstrates high consistency values. }
  \label{fig:motivation}
\end{wrapfigure}

Most black-box approaches probe a \emph{semantic equivalence space} by applying meaning-preserving perturbations to images or prompts and measuring response inconsistency~\cite{manakul2023selfcheckgpt, li2024reference, zhang2024vl}. For example, SelfCheckGPT samples multiple responses to the same query and scores their agreement using BERTScore, NLI entailment, unigram likelihood, or QA-based consistency~\cite{manakul2023selfcheckgpt}, while VL-Uncertainty perturbs the image and the question and measures the entropy over semantically clustered responses to estimate uncertainty~\cite{zhang2024vl}. These strategies reduce model behavior to a single inconsistency score, assuming that this one metric is sufficient to capture uncertainty to detect hallucination. However, we observe that LVLMs can generate hallucinated answers with low inconsistency scores on these metrics~\cite{kim2025medical, zhang2025siren} (see Figure~\ref{fig:motivation}), suggesting that uncertainty cannot be fully captured by a single metric.

Since uncertainty remains a valid and informative signal for hallucination detection~\cite{manakul2023selfcheckgpt}, the central challenge is not whether uncertainty should be measured, but how it manifests in vision-language models. Hallucinations do not necessarily exhibit a single, universal form of uncertainty. Some remain stable under textual resampling yet become inconsistent when the visual input is perturbed, while others appear robust to image transformations but fail under logically equivalent reasoning. Consequently, the uncertainty is inherently multifaceted, and reducing it to a single scalar score inevitably overlooks important behavioral evidence. Existing black-box methods, however, probe only one dimension of uncertainty, for example, by measuring inconsistency across textual resampling~\cite{li2024reference} or by aggregating image and question perturbations into a single entropy estimate~\cite{zhang2024vl}.

We therefore argue that hallucination detection should shift its focus from individual uncertainty metrics to the \emph{patterns} formed by multiple complementary manifestations of uncertainty. To this end, we retain the perturbation-modality axis explored in prior work, image transformations versus textual paraphrases, and introduce a second axis, \emph{logical polarity}, which probes a model with both an affirmative statement and its logical contradiction. This axis exploits a fundamental consistency constraint: if a model judges a statement about an image to be true, it should judge its logical negation to be false. By intersecting these two axes, our framework partitions model behavior into four complementary consistency patterns, each capturing a distinct manifestation of uncertainty. Rather than collapsing uncertainty into a single undifferentiated score, our approach models its structure, enabling us to analyze how different uncertainty patterns relate to hallucination and how much each contributes to accurate detection.

Based on this design, we propose \textbf{Unique Hallucination Pattern (UHP) Detection}, a fully black-box framework that extracts multiple within-group and between-group features from the defined four consistency groups to estimate uncertainty and trains a lightweight classifier to predict hallucination. The method requires only model outputs and is therefore applicable to both open- and closed-source LVLMs.

Through extensive experiments on AMBER~\cite{wang2023amber} and PhD~\cite{liu2025phd} using three LVLMs (InstructBLIP-7B\cite{dai2023instructblip}, InternVL-4B\cite{chen2024internvl}, and Qwen2.5-VL\cite{bai2025qwen2}), UHP Detection consistently outperforms prior state-of-the-art black-box methods, achieving up to \textbf{+18.72\%} and \textbf{+20.07\%} absolute improvements in \textbf{AUC-ROC} and \textbf{AUC-PR}, respectively, on AMBER, and up to \textbf{+13.26\%} and \textbf{+16.55\%} on PhD. 

Comprehensive ablation studies further substantiate the proposed structured consistency formulation. By isolating individual consistency groups and disentangling within-group from between-group features, we demonstrate that each group provides complementary signals for hallucination detection and that both feature types are essential for effective discrimination. Cross-dataset evaluations demonstrate generalizability, with classifiers trained on one benchmark retaining competitive performance when evaluated on another. This evidence confirms that hallucinations manifest as distinctive, model-dependent behavioral patterns within the proposed consistency space, suggesting that the learned representation captures transferable structures rather than overfitting to dataset-specific artifacts.

\paragraph{Contributions.}
We summarize our contributions as follows:
\begin{itemize}
    \item We introduce \textbf{UHP Detection}, a fully black-box hallucination detection framework applicable to both open- and closed-source LVLMs.
    
    \item We demonstrate consistent and substantial improvements over both white-box and black-box baselines across two benchmarks and three LVLMs.
    
    \item We propose a structured consistency space defined by perturbation modality and statement logical polarity, showing that hallucination manifests as a structured pattern across four consistency groups, validated through extensive ablation studies.
\end{itemize}

\section{Related Work}

\subsection{Large Vision–Language Models (LVLMs)}

LVLMs combine image encoders with large language models (LLMs) to support joint reasoning over visual and textual inputs. In typical designs, a visual backbone extracts image features, which are then transformed into “visual tokens” via a vision–language alignment module (e.g., MLPs, adapters, or Q-Formers), and finally fed into an LLM decoder for natural language output \cite{chen2023minigpt, liu2024improved, liu2023visual, dai2023instructblip, li2023blip, gao2023llama, touvron2023llama, chiang2023vicuna, bai2025qwen2}. Early LVLMs primarily targeted tasks such as captioning or VQA under global image-level prompts \cite{liu2024improved, liu2023visual, zhu2023minigpt}. Subsequent work extended LVLM capabilities to broader application domains including robotics, medical diagnosis, and autonomous driving \cite{huang2023embodied, peng2023kosmos, li2023llava, moor2023med, wang2023drivemlm, cui2024survey}. Despite impressive multimodal performance, LVLMs remain prone to generating content that does not faithfully reflect the visual input called hallucinations.

\subsection{Hallucinations in LVLMs}
Despite impressive multimodal performance, LVLMs remain prone to generating content that does not faithfully reflect the visual input called hallucinations. Hallucinations in LVLMs commonly categorized into object existence, attribute misdescription, and relational errors~\cite{yin2024survey}. Numerous benchmarks have been proposed to systematically evaluate these categories. Early efforts like POPE~\cite{li2023evaluating} probe object hallucinations through targeted yes/no questions on image content. Recent taxonomies extend this framework to include more diverse and subtle forms of hallucinations, such as multi-modal conflicts and counter-common-sense failures~\cite{rawte2025defining, sun2025understanding, liu2025phd}. AMBER~\cite{wang2023amber} offers a multi-dimensional, LLM-free assessment across various hallucination types. Moreover, the recent PhD benchmark~\cite{liu2025phd} introduces a comprehensive taxonomy of hallucinations, categorizing them into object-, attribute-, multimodal-conflict-, and counter–common-sense-based errors. This taxonomy covers a wide range of visual understanding tasks, from low-level perception such as object and attribute recognition to mid-level reasoning tasks including spatial relationship understanding and object counting.

\subsection{Hallucination Detection in LVLMs}
 
Most early attempts at detecting hallucinations in LVLMs rely on reference-based signals, such as external detectors~\cite{chen2024unified, gao2024aigcs, yin2024woodpecker} or ground-truth annotations~\cite{wang2023amber, li2023evaluating}. While these methods can provide high fidelity under controlled conditions, their practical adoption is limited because they require access to reliable external models or human annotations~\cite{li2024reference}. To overcome these shortcomings, recent research has shifted toward \emph{reference-free} detection methods that exploit intrinsic signals from the LVLM itself. Specifically, uncertainty-based detection has emerged as a promising paradigm~\cite{fadeeva2023lm, li2024reference, lin2023generating, manakul2023selfcheckgpt, zhang2024vl}. 

Uncertainty-based hallucination detection methods can be broadly categorized by the level of access they require to the underlying model~\cite{bai2024hallucination}. \textbf{White-box methods} leverage internal model signals, such as token-level probabilities, hidden representations, gradients, or attention patterns to define their uncertainty metrics~\cite{zhao2025visual, zollicoffer2025diverging, duan2024shifting, zhao2024first, guerreiro2023looking, kadavath2022language}. In contrast, \textbf{black-box methods} treat the model as an opaque system and rely solely on observable input–output behavior, making them applicable to closed-source models. A common strategy is to sample multiple outputs for the same input, or for semantically equivalent perturbations, and quantify the inconsistency among the generated responses. High output variability is then used as a metric for model uncertainty and potential hallucination~\cite{li2024reference, gautam2025hedge, manakul2023selfcheckgpt, zhang2024vl}. 

However, recent studies show that LVLMs can produce hallucinated answers with low uncertainty or inconsistency metrics~\cite{kim2025medical, zhang2025siren}, which limits the effectiveness of existing black-box methods. Instead of reducing all generated samples to a single uncertainty score, our approach introduces a structured consistency space exploration based on two complementary axes. These two axes, perturbation source and logical polarity, define the sampling methods, which are organized into four consistency groups. Based on these groups, we compute both within-group and between-group consistency features. This design enables a more structured characterization of model behavior and reveals consistency patterns that simpler metrics may overlook.

\begin{figure}[t]
  \centering
  \includegraphics[width=\linewidth]{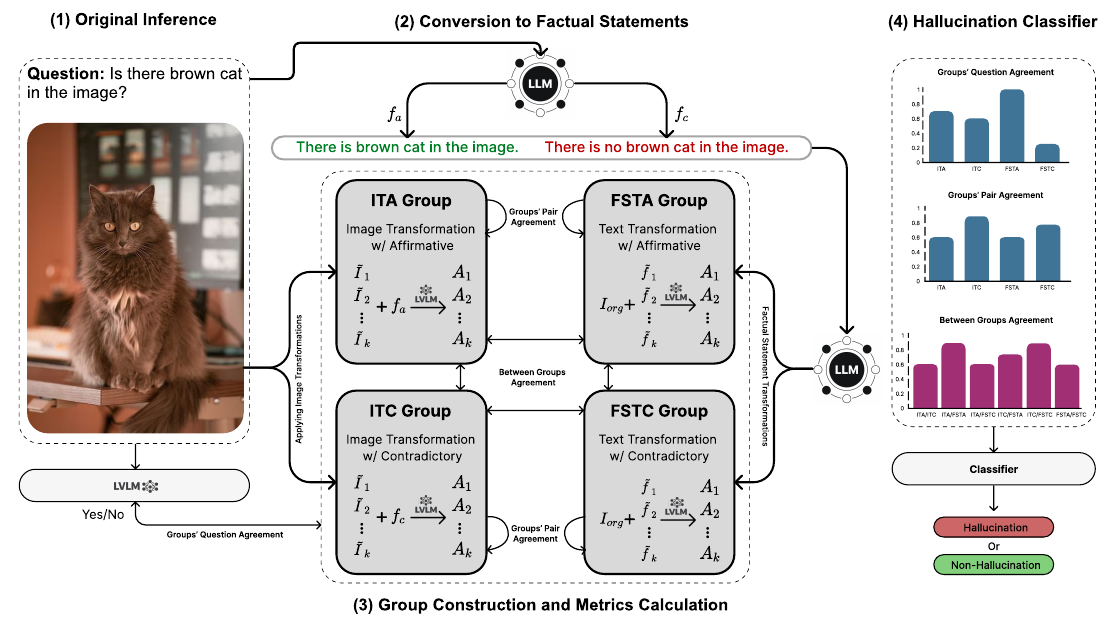}
  \caption{Overview of our framework. (1) \textbf{Original Inference}: The LVLM generates a preliminary answer for the given image-question pair. (2) \textbf{Conversion to Factual Statement }: The question is transformed into affirmative ($f_a$) and contradictory ($f_c$) statements to probe logical polarity. (3) \textbf{Group Construction and Metrics Calculation}: We construct four distinct consistency groups by applying semantic-equivalent perturbations to both images and statements. For each group, we calculate within-group and between-group features (Agreement metrics) to capture behavioral inconsistencies. (4) \textbf{Hallucination Classifier}: A lightweight classifier processes the extracted 14-dimensional feature vector to predict whether the original response is a hallucination.}
  \label{fig:mainfigure}
\end{figure}

\section{Method}

\subsection{Problem Formulation and Notation}

Let \(V\) be a LVLM. Given an image \(I\) and a binary question \(Q\) drawn directly from the dataset, the model generates a response sequence \(r = (y_1, \dots, y_L)\) conditioned on the image–question pair \((I, Q)\). Since we focus on binary questions, the generated response \(r\) is mapped to \(\hat{a} \in \mathcal{A} = \{\texttt{0}, \texttt{1}\}\), where \(\texttt{0}\) denotes ``no'' and \(\texttt{1}\) denotes ``yes''. For each input pair \((I, Q)\), the dataset provides a ground-truth label \(a^\ast \in \mathcal{A}\), indicating whether the queried content is correct with respect to the image. A \emph{hallucination} occurs when the model’s prediction disagrees with the ground truth (\(\hat{a} \neq a^\ast\)). In this work, our objective is to detect such hallucinations in the responses generated by \(V\).

\subsection{Consistency Space Exploration}

We propose a black-box hallucination detection framework that explores the \emph{consistency space} of LVLMs. The central idea is that hallucinations can be revealed through structured inconsistencies in model behavior. To characterize this space, we examine two complementary axes: (i) \emph{logical polarity} and (ii) \emph{source of perturbation} (see Figure~\ref{fig:mainfigure}).

\subsubsection{Logical Polarity}

We investigate logical polarity based on a fundamental consistency principle: if an LVLM judges a statement $A$ about an image to be true, it should judge its negation $\neg A$ to be false. To implement this idea, we transform a base question $Q$ into two factual statements: an affirmative statement ($f_a$) and a contradictory statement ($f_c$). These statements are generated using an LLM with predefined prompts (see Appendix A.1) under controlled assumptions about the answer to $Q$. Specifically, $f_a$ is produced by assuming the answer to $Q$ is always “Yes,” while $f_c$ is generated by assuming the answer is always “No.” For example, given the question $Q=$ “Is there a cat in the image?”, the generated statements would be $f_a=$ “There is a cat in the image” and $f_c=$ “There is no cat in the image.” These paired statements allow us to probe cross-polarity logical coherence.

\subsubsection{Source of Perturbation}

The second axis explores the source of perturbation. A reliable model should remain consistent across semantically equivalent inputs, samples that differ in representation but preserve the same underlying meaning. We therefore construct controlled perturbations along both the visual and textual modalities.

\paragraph{Image-Based Transformations}

Given an image $I$, we construct a set of semantically equivalent images $\mathcal{S}(I)=\{\tilde{I}_k\}_{k=1}^{N}$ by applying carefully selected image transformations. The key criterion is that these transformations must not alter the factual content of the image. To ensure semantic equivalence, candidate transformations are evaluated using the scene composition similarity metric proposed in~\cite{haque2025novel}, which measures structural and compositional similarity between the original image $I$ and each transformed image $\tilde{I}_k$. Only transformations with high similarity scores are retained. Detailed similarity statistics and qualitative examples are provided in the Appendix B.

\paragraph{Factual Statement Transformations}

For each factual statement $f$, we generate a set of semantically equivalent variants $\mathcal{S}(f)=\{\tilde{f}_k\}_{k=1}^{M}$, covering both affirmative and contradictory forms ($f_a$, $f_c$). Paraphrases are generated using an LLM guided by handcrafted templates and in-context examples to minimize semantic drift. The templates implement four controlled paraphrasing operations:  
(i) \emph{lexical substitution}, replacing words or short phrases with synonyms;  
(ii) \emph{syntactic restructuring}, modifying sentence structure (e.g., inversion or cleft constructions);  
(iii) \emph{phrasal rephrasing}, substituting multi-word expressions with equivalent alternatives; and  
(iv) \emph{addition of neutral modifiers}, introducing non-informative descriptors (e.g., “clearly,” “visible”) that preserve meaning.

Finally, each factual statement is embedded into a templated prompt (see Appendix A.2) and provided to the model together with the corresponding image. The model is required to evaluate the statement as either \textit{True} or \textit{False}, enabling systematic consistency analysis across perturbations and logical polarities.

\subsection{Consistency Group Definition}  

By intersecting the two types of perturbation (visual vs. textual) with the two logical forms of factual statements (affirmative vs. contradictory), we establish four distinct consistency groups.

\paragraph{Image-based Consistency Groups}
In these groups, we fix the textual input and measure consistency across the set of semantically equivalent images $\mathcal{S}(I)$.
\begin{itemize}
    \item Image Transformation w/ Affirmative (ITA): The model evaluates the fixed affirmative statement $f_a$ against each transformed image $\tilde{I}_k \in \mathcal{S}(I)$.
    \item Image Transformation w/ Contradictory (ITC): The model evaluates the fixed contradictory statement $f_c$ against each transformed image $\tilde{I}_k \in \mathcal{S}(I)$.
\end{itemize}

\paragraph{Text-based Consistency Groups}
In these groups, we fix the original image $I$ and measure consistency across the set of semantically equivalent factual statements $\mathcal{S}(f)$.
\begin{itemize}
    \item Factual Statement Transformation w/ Affirmative (FSTA): The model evaluates each paraphrased affirmative statement $\tilde{f}_k \in \mathcal{S}(f_a)$ against the fixed original image $I$. 
    \item Factual Statement Transformation w/ Contradictory (FSTC): The model evaluates each paraphrased contradictory statement $\tilde{f}_k \in \mathcal{S}(f_c)$ against the fixed original image $I$.
\end{itemize}

\subsection{Consistency Metrics \& Hallucination Detection}

For each consistency group \(G \in \{ITA, ITC, FSTA, FSTC\}\), we collect a set of binary predictions
$
\mathcal{Y}_G = \{ y_1, \dots, y_{|G|} \}, \quad \forall y_i \in \{0, 1\}, 
$
where \(0\) denotes \emph{False} and \(1\) denotes \emph{True} and \(|G| = N\) for image-based and \(|G| = M\) for text-based consistency groups. We define two within metrics: (i) \emph{question-aligned agreement} and (ii) \emph{pairwise agreement}, which together serve as uncertainty signals for hallucination detection.

\paragraph{Question-aligned agreement}
This metric measures how much predictions within a consistency group $\mathcal{Y}_G$ agree with the model’s original answer \(\hat{a}\). Since affirmative and contradictory groups have opposite logical semantics, we define question-aligned agreement as
\begin{equation}
\mathrm{QA}(G) =
\begin{cases}
\displaystyle \frac{1}{|G|} \sum_{y \in \mathcal{Y}_G} \mathbb{I}\!\left[y = \hat{a}\right],
& G \in \{\mathrm{ITA}, \mathrm{FSTA}\} \\[6pt]
\displaystyle \frac{1}{|G|} \sum_{y \in \mathcal{Y}_G} \mathbb{I}\!\left[y \neq \hat{a}\right],
& G \in \{\mathrm{ITC}, \mathrm{FSTC}\}
\end{cases}.
\label{eq:qa}
\end{equation}

\paragraph{Pairwise agreement}
Pairwise agreement measures the consistency of predictions within a consistency group by quantifying how often pairs of predictions agree with each other. For a consistency group \(G\), the pairwise agreement is computed as
\begin{equation}
\mathrm{PA}(G) = \frac{1}{\binom{|G|}{2}} \sum_{i<j} \mathbb{I}[y_i = y_j].
\end{equation}

\paragraph{Between-group features}
In addition to within-group metrics, we construct six between-group features based on absolute differences between question-aligned agreement scores. These features measure the degree of deviation between groups in the four consistency groups, indicating how much two groups agree. For selected pairs of groups \((G_i, G_j)\), we define
\begin{equation}
\mathrm{BG}(G_i, G_j) = 
\left| \mathrm{QA}(G_i) - \mathrm{QA}(G_j) \right|.
\end{equation}

\paragraph{Feature construction and classifier}
For each sample, we compute \(\mathrm{QA}(G)\) and \(\mathrm{PA}(G)\) for each of the four consistency groups, resulting in eight within-group features. We also compute six between-group agreement features and Altogether, this yields fourteen scalar consistency features per sample. Finally, we train a lightweight binary detector \(C_{\psi}\) on the 14-dimensional consistency features.

Experiment Sections
\section{Experiments}
\subsection{Experimental setup}

\paragraph{Datasets \& LVLMs}
We evaluate our method on two challenging benchmarks for visual hallucination: AMBER and PhD~\cite{wang2023amber, liu2025phd}. These datasets are selected over earlier, saturated benchmarks like POPE as they provide a broader and more rigorous evaluation scope~\cite{li2024reference, zhao2025visual}. We assess performance across three diverse open-source LVLMs: InstructBLIP-7B~\cite{dai2023instructblip}, Qwen2.5-VL~\cite{bai2025qwen2}, and InternVL-4B~\cite{chen2024internvl}. These models are chosen for their distinct architectural designs and competitive multimodal capabilities, while their open-source nature facilitates direct comparison with existing white-box approaches.


\paragraph{Baselines}
We compare our method with a set of state-of-the-art  white-box and black-box hallucination detection approaches. For white-box methods, we include AvgProb, AvgEnt, MaxProb, MaxEnt, P\_True, SeqLogProb, and TokenSAR~\cite{li2024reference, duan2024shifting, guerreiro2023looking, kadavath2022language}. For black-box methods, we consider Question Answering (QA), BERTScore, Unigram, NLI, and VL-Uncertainty~\cite{li2024reference, zhang2024vl}.


\paragraph{Implementation Details}

For each dataset, we randomly sample 2,000 instances and perform stratified sampling to preserve an equal distribution of ``yes'' and ``no'' labels across all splits. The data is divided into 80\% training/validation and 20\% test sets. Classifier selection is performed exclusively on the training/validation split using the AUC-ROC metric. We use AUC-ROC for model selection because it is threshold-independent and robust to class imbalance; consistent trends are also observed with AUC-PR, which we report alongside all results.

To generate the required factual statements, we employ Phi-2~\cite{gunasekar2023textbooksneed} as the underlying LLM. Specifically, Phi-2 is used to (i) convert binary questions into affirmative and contradictory factual statements and (ii) generate semantically equivalent paraphrases of these statements. For the visual modality, we apply 10 semantic-preserving image transformations (Appendix~B).

Using these perturbations, we construct the four consistency groups and extract the 14-dimensional consistency features for each sample. We evaluate a pool of eight classifiers (Appendix~C), including both linear and non-linear models, and select the best-performing classifier based on validation AUC-ROC. To ensure a fair comparison, the same classifier pool, training protocol, and validation-based selection procedure are applied to all baseline methods. Specifically, we compute each baseline's  metric using its original implementation and treat the classifier as a common post-processing component. This protocol avoids favoring any particular method while providing a consistent evaluation framework across both training-based and training-free approaches.

\subsection{Main Results}

Tables~\ref{tab:results-AMBER} and~\ref{tab:results-PhD} report the hallucination detection performance of UHP Detection against state-of-the-art white-box and black-box baselines across the AMBER and PhD benchmarks, utilizing three distinct LVLM architectures: InstructBlip-7B, Qwen2.5-VL, and InternVL-4B. The evaluation metrics include Accuracy, AUC-ROC, and AUC-PR.

On the AMBER dataset, our method establishes a consistent performance advantage, achieving the highest scores across all metrics and model architectures. Specifically, our method attains an AUC-PR of 70.10\% on Qwen2.5-VL, substantially outperforming the strongest baseline, Unigram, which scores 50.07\%. This trend is replicated on InstructBlip-7B and InternVL-4B, where UHP Detection records AUC-PR scores of 54.59\% and 49.80\%, respectively, compared to the best baselines of 37.78\% and 36.81\%.

On the PhD dataset, the overall performance of all methods declines, highlighting the complexity of the benchmark. Despite this increased difficulty, our method maintains its superiority in nearly all cases. Notably, on Qwen2.5-VL, our method achieves an AUC-PR of 53.36\%, significantly surpassing the second-best method, AvgEnt (47.37\%). However, the results reveal a notable competitive gap on InstructBlip-7B, where the white-box method AvgEnt slightly edges out our method in AUC-PR (45.87\% vs. 44.28\%). Furthermore, the results underscore the instability of existing baselines, as evidenced by the P\_True method, which achieves the high accuracy on InternVL-4B in the PHD dataset (74.25\%) but suffers from a critical collapse in AUC-PR (26.20\%), whereas UHP Detection maintains more robust balance between precision and recall than other methods across all scenarios.

\begin{table*}[!ht]
\centering
\setlength{\tabcolsep}{4pt}
\renewcommand{\arraystretch}{1.15}
\resizebox{\textwidth}{!}{
\begin{tabular}{ll ccc ccc ccc}
\toprule
\multirow{2}{*}{\textbf{Type}} & \multirow{2}{*}{\textbf{Method}} 
& \multicolumn{3}{c}{\textbf{InstructBlip-7B}} 
& \multicolumn{3}{c}{\textbf{Qwen2.5-VL}} 
& \multicolumn{3}{c}{\textbf{InternVL-4B}} \\
\cmidrule(lr){3-5} \cmidrule(lr){6-8} \cmidrule(lr){9-11}
& & ACC & AUC-ROC & AUC-PR & ACC & AUC-ROC & AUC-PR & ACC & AUC-ROC & AUC-PR \\
\midrule

\multirow{7}{*}{\rotatebox{90}{White-box}}
& AvgProb    & 60.75 & 64.23 & \underline{37.78} & 68.75 & 73.91 & 47.55 & 68.50 & 67.19 & 27.02 \\
& AvgEnt     & 58.75 & 61.80 & 32.17 & 74.25 & 75.37 & 45.96 & 69.50 & 66.81 & 29.73 \\
& MaxProb    & 66.25 & 60.31 & 33.36 & 73.75 & 73.83 & 47.20 & 71.25 & \underline{70.45} & 35.20 \\
& MaxEnt     & 60.25 & 60.36 & 32.57 & 65.50 & 71.22 & 42.56 & 67.00 & 66.99 & \underline{36.81} \\
& P\_True    & 47.21 & 55.58 & 26.93 & 67.82 & 58.45 & 21.61 & \underline{72.75} & 63.41 & 27.44 \\
& SeqLogProb & 57.98 & 64.43 & 35.12 & 67.26 & 72.48 & 37.72 & 69.25 & 64.45 & 23.76 \\
& TokenSAR   & 65.42 & \underline{67.99} & 37.45 & 66.98 & 70.85 & 35.48 & 57.00 & 59.83 & 24.79 \\

\midrule
\multirow{5}{*}{\rotatebox{90}{Black-box}}
& QA           & 58.75 & 58.18 & 30.53 & 61.50 & 65.48 & 38.73 & 60.50 & 51.42 & 26.16 \\
& BertScore    & 65.00 & 58.97 & 33.34 & 71.75 & 73.33 & 44.96 & 64.50 & 53.51 & 28.35 \\
& Unigram      & 54.50 & 59.09 & 30.76 & \underline{77.25} & \underline{76.06} & \underline{50.07} & 57.75 & 57.95 & 29.93 \\
& NLI          & \underline{68.00} & 60.41 & 34.52 & 64.25 & 59.25 & 33.51 & 68.50 & 59.78 & 32.81 \\
& vl-uncertainty & 58.25 & 54.67 & 29.13 & 66.50 & 64.08 & 25.02 & 65.75 & 62.57 & 21.52 \\

\rowcolor{lightblue}
& \textbf{UHP Detection}
& \textbf{77.40} & \textbf{78.03} & \textbf{54.59}
& \textbf{85.89} & \textbf{84.47} & \textbf{70.10}
& \textbf{84.75} & \textbf{81.29} & \textbf{49.80} \\
\bottomrule
\end{tabular}}
\caption{Hallucination detection performance on AMBER. Best results are bold; second-best are underlined.}
\label{tab:results-AMBER}
\end{table*}

\begin{table}[!h]
\centering
\setlength{\tabcolsep}{4pt}
\renewcommand{\arraystretch}{1.15}
\resizebox{\textwidth}{!}{
\begin{tabular}{ll ccc ccc ccc}
\toprule
\multirow{2}{*}{\textbf{Type}} & \multirow{2}{*}{\textbf{Method}} 
& \multicolumn{3}{c}{\textbf{InstructBlip-7B}} 
& \multicolumn{3}{c}{\textbf{Qwen2.5-VL}} 
& \multicolumn{3}{c}{\textbf{InternVL-4B}} \\
\cmidrule(lr){3-5} \cmidrule(lr){6-8} \cmidrule(lr){9-11}
& & ACC & AUC-ROC & AUC-PR & ACC & AUC-ROC & AUC-PR & ACC & AUC-ROC & AUC-PR \\
\midrule
\multirow{7}{*}{\rotatebox{90}{White-box}}
& AvgProb  & 64.62 & 63.01 & \underline{45.36} & 73.75 & \underline{74.49} & 43.77 & 56.50 & 52.94 & 27.43 \\
& AvgEnt   & \underline{66.00} & \underline{63.16} & \textbf{45.87} & 74.25 & 73.58 & \underline{47.37} & 62.50 & 60.08 & 34.58 \\
& MaxProb  & 62.87 & 61.36 & 41.03 & 68.25 & 72.26 & 43.59 & 65.25 & 56.82 & 29.02 \\
& MaxEnt   & 65.87 & \textbf{63.25} & 43.29 & 74.22 & 71.22 & 42.56 & 56.50 & 53.62 & 28.69 \\
& P\_True  & 40.20 & 53.98 & 32.93 & 67.82 & 58.45 & 21.61 & \underline{74.25} & 55.81 & 26.20 \\
& SeqLogProb & 59.35 & 57.72 & 34.34 & 68.67 & 72.86 & 37.72 & 59.37 & 56.83 & 34.26 \\
& TokenSAR & 60.22 & 62.74 & 40.68 & 66.98 & 70.85 & 35.48 & 51.04 & 51.07 & 26.44 \\
\midrule
\multirow{5}{*}{\rotatebox{90}{Black-box}}
& QA         & 62.23 & 56.54 & 37.96 & 62.23 & 64.48 & 33.08 & 66.40 & 54.90 & 30.34 \\
& BertScore  & 60.93 & 58.04 & 40.22 & 64.84 & 70.67 & 36.81 & 58.59 & 57.09 & 34.38 \\
& Unigram    & 58.07 & 59.46 & 41.04 & 65.62 & 65.86 & 33.96 & 61.45 & 56.63 & 33.50 \\
& NLI        & 64.04 & 61.11 & 40.38 & \underline{77.86} & 63.61 & 34.04 & 64.06 & 64.14 & \underline{36.40} \\
& vl-uncertainty & 57.50 & 58.75 & 38.88 & 66.25 & 61.83 & 24.25 & 68.75 & \underline{69.02} & 35.38 \\
\rowcolor{lightblue}
& \textbf{UHP Detection}
& \textbf{68.73} & 60.45 & 44.28
& \textbf{81.25} & \textbf{79.12} & \textbf{53.36}
& \textbf{78.98} & \textbf{72.29} & \textbf{45.86} \\
\bottomrule
\end{tabular}}
\caption{Hallucination detection performance on PhD. Best results are bold; second-best are underlined.}
\label{tab:results-PhD}
\end{table}

\subsection{Ablation on Consistency Group Contributions}
To examine the contribution of each consistency group, we conduct controlled ablations by selectively removing individual groups (ITA, ITC, FSTA, FSTC), restricting to single-modality perturbations subsets (image-only or text-only), and isolating logical polarity (affirmative-only or contradictory-only). Results are shown in Appendix D.

The results show that no individual group matches the performance of the full UHP Detection configuration, as removing any component results in a consistent AUC-ROC drop of 3–9 \% on AMBER and a more clear decline on PhD. Regarding modality, restricting our method to image-only or text-only subsets consistently underperforms the full configuration, with gaps reaching up to 9\% AUC-ROC on PhD, which demonstrates that visual perturbation and linguistic paraphrase consistencies encode complementary, non-redundant signals. 

Furthermore, logical polarity modeling highlights the importance of cross-polarity coherence; while affirmative-only and contradictory-only groups offer meaningful detection, their combination yields superior results, indicating that hallucinations involve violations of logical consistency between statements and their negations. Ultimately, the full UHP Detection configuration achieves the highest performance by integrating cross-modality consistency signals and cross-polarity coherence, empirically supporting the hypothesis that hallucinations manifest as structured patterns in the consistency space.

\subsection{Impact of Within- and Between-Group Agreement Metrics}

To quantify the contribution of within- and between-group agreement metrics, we ablate the feature space along three axes: (i) using only within-group agreement features (QA and PA computed per each consistency group), (ii) using only between-group agreement features, and (iii) the full 14 features combining both. As reported in Table~\ref{tab:merged_within_between}, within-group features consistently serve as the primary signal for hallucination detection, outperforming between-group features across all metrics and model architectures. 

For instance, on the AMBER dataset, the within-group model achieves an AUC-PR of 66.07\% for Qwen2.5-VL, whereas the between-group counterpart lags at 54.82\%. However, relying solely on within-group consistency is insufficient for optimal performance. The integration of between-group agreement features yields a synergistic effect, boosting detection capabilities. For example, in the AMBER dataset, where adding between-group features to the within-group baseline improves the AUC-PR on InstructBlip-7B from 40.81\% to 54.59\%.

\begin{table*}[!h]
\centering
\resizebox{\textwidth}{!}{%
\begin{tabular}{>{\raggedright\arraybackslash}p{1.4cm} l ccc ccc ccc}
\toprule
\multirow{2}{*}{\textbf{Dataset}} & \multirow{2}{*}{\textbf{Method}}
& \multicolumn{3}{c}{\textbf{InstructBlip-7B}}
& \multicolumn{3}{c}{\textbf{QWEN2.5-VL}}
& \multicolumn{3}{c}{\textbf{InternVL-4B}} \\
\cmidrule(lr){3-5} \cmidrule(lr){6-8} \cmidrule(lr){9-11}
& & \textbf{ACC} & \textbf{AUC-ROC} & \textbf{AUC-PR}
& \textbf{ACC} & \textbf{AUC-ROC} & \textbf{AUC-PR}
& \textbf{ACC} & \textbf{AUC-ROC} & \textbf{AUC-PR} \\
\midrule
\multirow{3}{*}{AMBER}
& Within-group only  & 68.00 & 62.49 & 40.81 & 84.48 & 81.52 & 66.07 & 81.00 & 77.76 & 49.22 \\
& Between-group only & 55.25 & 57.94 & 37.73 & 74.10 & 74.45 & 54.82 & 77.75 & 78.32 & 43.90 \\
& \textbf{UHP Detection} & \textbf{77.40} & \textbf{78.03} & \textbf{54.59} & \textbf{85.89} & \textbf{84.47} & \textbf{70.10} & \textbf{84.75} & \textbf{81.29} & \textbf{49.80} \\
\midrule
\multirow{3}{*}{PhD}
& Within-group only  & 61.50 & 59.43 & 40.20 & 76.50 & 78.40 & 51.13 & 65.50 & 68.12 & 41.54 \\
& Between-group only & 50.25 & 55.51 & 34.48 & 64.25 & 65.56 & 34.06 & 60.50 & 59.21 & 28.43 \\
& \textbf{UHP Detection} & \textbf{68.73} & \textbf{60.45} & \textbf{44.28} & \textbf{81.25} & \textbf{79.12} & \textbf{53.36} & \textbf{78.98} & \textbf{72.29} & \textbf{45.86} \\
\bottomrule
\end{tabular}}
\caption{\textbf{Impact of within- and between-group agreement on AMBER and PhD.}}
\label{tab:merged_within_between}
\end{table*}

\subsection{Sensitivity to Perturbation Sampling Size}

We analyze the robustness of our method by varying the number of image (N) and text (M) transformations, as reported in Table~\ref{tab:sens_results}. The results demonstrate a clear saturation trend across all LVLMs. This is particularly evident in the AUC-ROC scores for QWEN2.5-VL, where performance exhibits noticeable fluctuations as the sampling size increases; for instance, the score rises from 81.66\% ($\mathcal{I}_{2}\text{-}\mathcal{T}_{2}$)  to 84.19\% ($\mathcal{I}_{8}\text{-}\mathcal{T}_{2}$), but oscillates in final steps and shows minimal further improvement beyond this point. This indicates that UHP Detection achieves its effectiveness through structured group-based consistency space exploration rather than brute-force sampling size expansion, ensuring computational efficiency and stability under reasonable transformation budgets.

\begin{table*}[!h]
\centering
\resizebox{\textwidth}{!}{%
\begin{tabular}{l c ccc ccc ccc}
\toprule
\multirow{2}{*}{\textbf{Method}} & \multirow{2}{*}{\textbf{\# Inf.}}
& \multicolumn{3}{c}{\textbf{InstructBlip-7B}}
& \multicolumn{3}{c}{\textbf{QWEN2.5-VL}}
& \multicolumn{3}{c}{\textbf{InternVL-4B}} \\
\cmidrule(lr){3-5} \cmidrule(lr){6-8} \cmidrule(lr){9-11}
&
& \textbf{ACC} & \textbf{AUC-ROC} & \textbf{AUC-PR}
& \textbf{ACC} & \textbf{AUC-ROC} & \textbf{AUC-PR}
& \textbf{ACC} & \textbf{AUC-ROC} & \textbf{AUC-PR} \\
\midrule
$\mathcal{I}_{2}\text{-}\mathcal{T}_{2}$      & 10  & 73.50 & 73.05 & 48.58 & 83.58 & 81.66 & 58.05 & 79.00 & 76.38 &  40.61 \\
$\mathcal{I}_{2}\text{-}\mathcal{T}_{\text{4}}$     & 14  & 74.25 & 72.05 & 49.51 & 83.33 & 81.00 & 58.15 & 79.75 & 77.29 & 41.91 \\
$\mathcal{I}_{4}\text{-}\mathcal{T}_{2}$      & 14  & 74.45 & 73.27 & 49.99 & 83.58 & 82.03 & 62.86 & 79.50 & 78.30 & 42.97 \\
$\mathcal{I}_{4}\text{-}\mathcal{T}_{4}$    & 18 & 75.50 & 73.78 & 50.32 & 82.82 & 82.19 & 65.98 & 80.25 & 79.20 & 43.72 \\
$\mathcal{I}_{6}\text{-}\mathcal{T}_{2}$       & 18 & 75.00 & 73.95 & 51.67 & 83.58 & 82.75 & 66.14 & 82.75 & 79.24 & 44.54 \\
$\mathcal{I}_{6}\text{-}\mathcal{T}_{4}$      & 22 & 75.75 & 74.45 & 52.35 & 84.10 & 83.76 & 66.36 & 83.50 & 79.15 & 46.56 \\
$\mathcal{I}_{8}\text{-}\mathcal{T}_{2}$     & 22 & 76.00 & 75.70 & 52.83 & 84.61 & 84.19 & 66.04 & 84.00 & 80.44 & 45.91 \\
$\mathcal{I}_{8}\text{-}\mathcal{T}_{4}$   & 26 & 76.00 & 75.90 & 52.02 & 85.38 & 83.98 & 67.51 & 83.75 & 79.99 & 46.71 \\
$\mathcal{I}_{10}\text{-}\mathcal{T}_{2}$    & 26 & 77.00 & 76.61 & 53.06 & 85.12 & \textbf{85.03} & 66.57 & \textbf{85.25} & 81.01 & \textbf{51.36} \\
$\mathcal{I}_{10}\text{-}\mathcal{T}_{4}$    & 30 & \textbf{77.40} & \textbf{78.03} & \textbf{54.59}
& \textbf{85.89} & 84.47 & \textbf{70.10}
& 84.75 & \textbf{81.29} & 49.80 \\
\bottomrule
\end{tabular}%
}
\caption{\textbf{Sensitivity to perturbation sampling size on the AMBER dataset.} $\mathcal{I}_{N}$ and $\mathcal{T}_{M}$ denote the number of image and text transformations, respectively. Best is bold.}
\label{tab:sens_results}
\end{table*}

\subsection{Hallucination Pattern in Consistency Space}

\begin{figure}[!h]
    \centering
    \begin{subfigure}[b]{0.45\textwidth}
        \centering
        \includegraphics[width=0.8\textwidth]{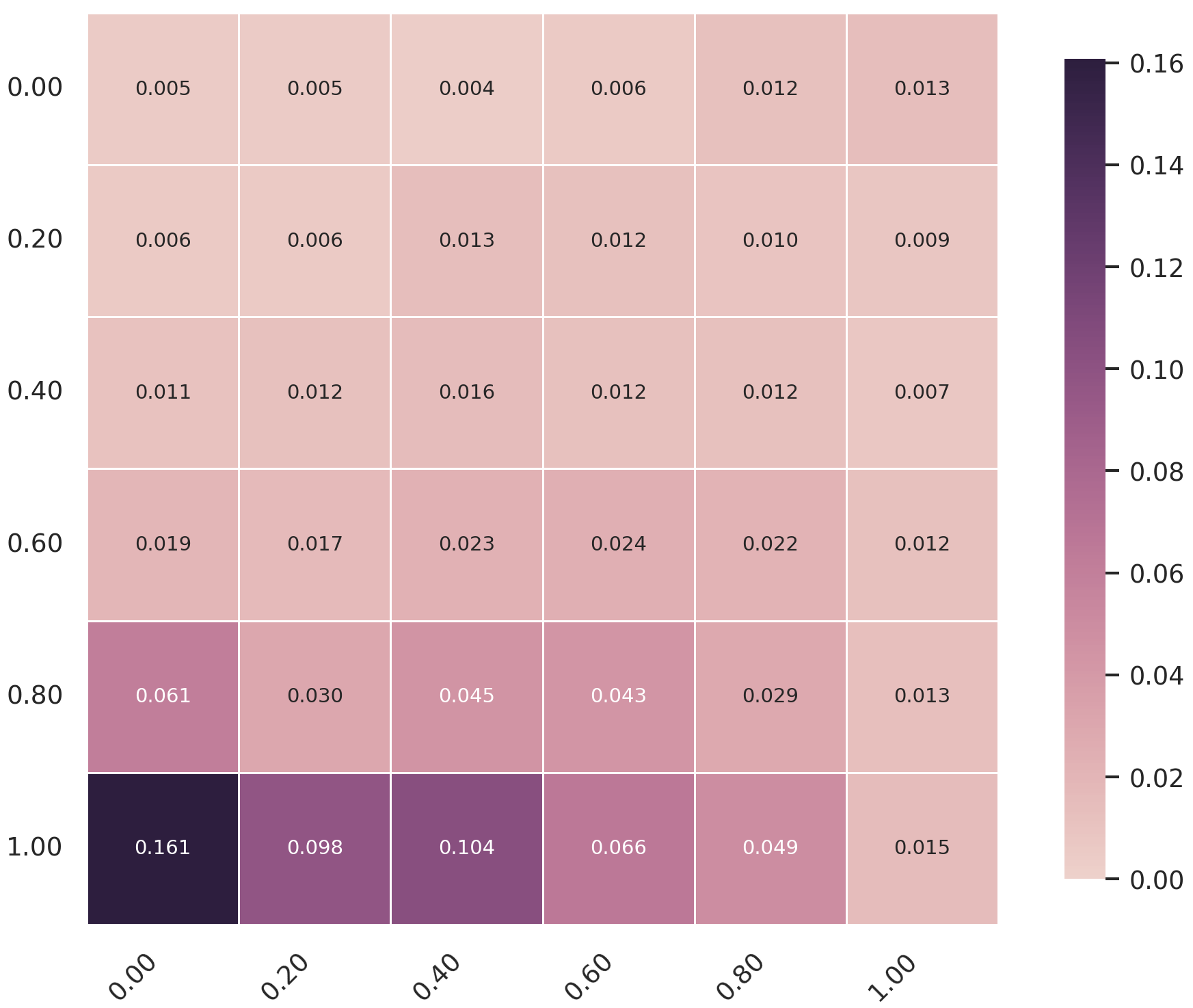} 
        \caption{FSTC vs. FSTA - Non hallucinations}
        \label{fig:plot1}
    \end{subfigure}
    \hfill
    \begin{subfigure}[b]{0.45\textwidth}
        \centering
        \includegraphics[width=0.8\textwidth]{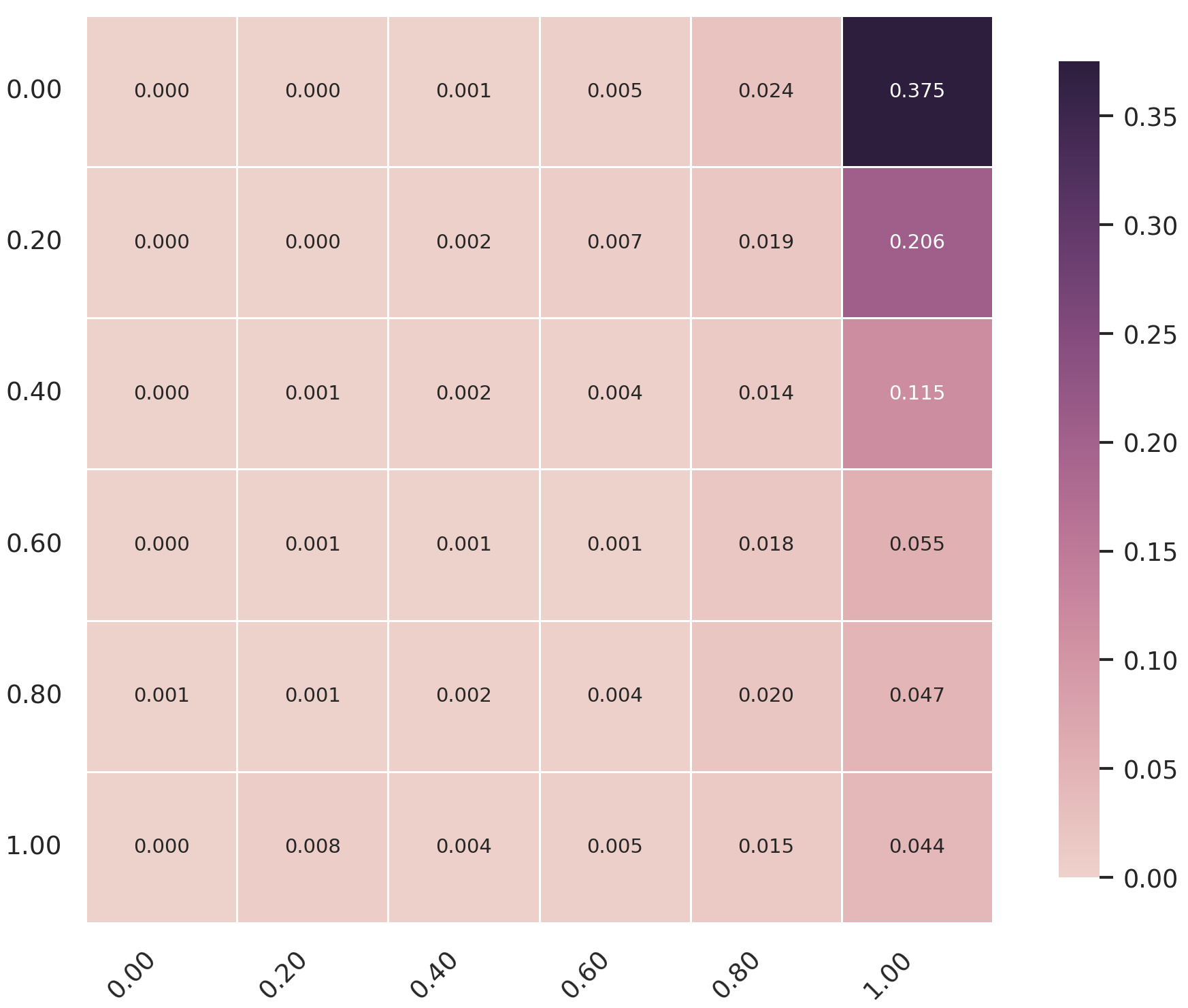} 
        \caption{FSTC vs. FSTA - Hallucinations}
        \label{fig:plot2}
    \end{subfigure}
    
    \vspace{0.1cm} 
    
    \begin{subfigure}[b]{0.45\textwidth}
        \centering
        \includegraphics[width=0.8\textwidth]{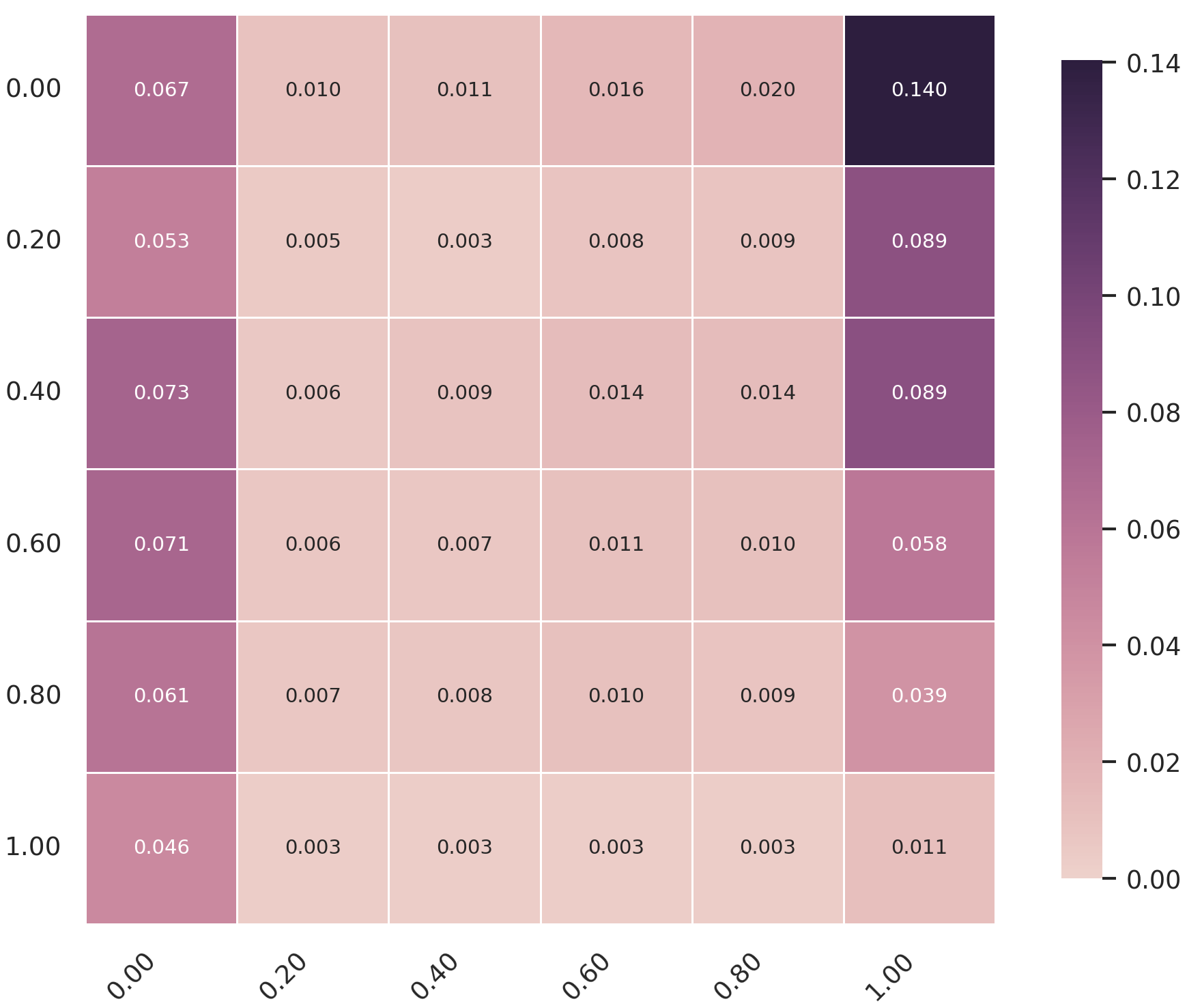} 
        \caption{FSTA vs. ITC - Non hallucinations}
        \label{fig:plot3}
    \end{subfigure}
    \hfill
    \begin{subfigure}[b]{0.45\textwidth}
        \centering
        \includegraphics[width=0.8\textwidth]{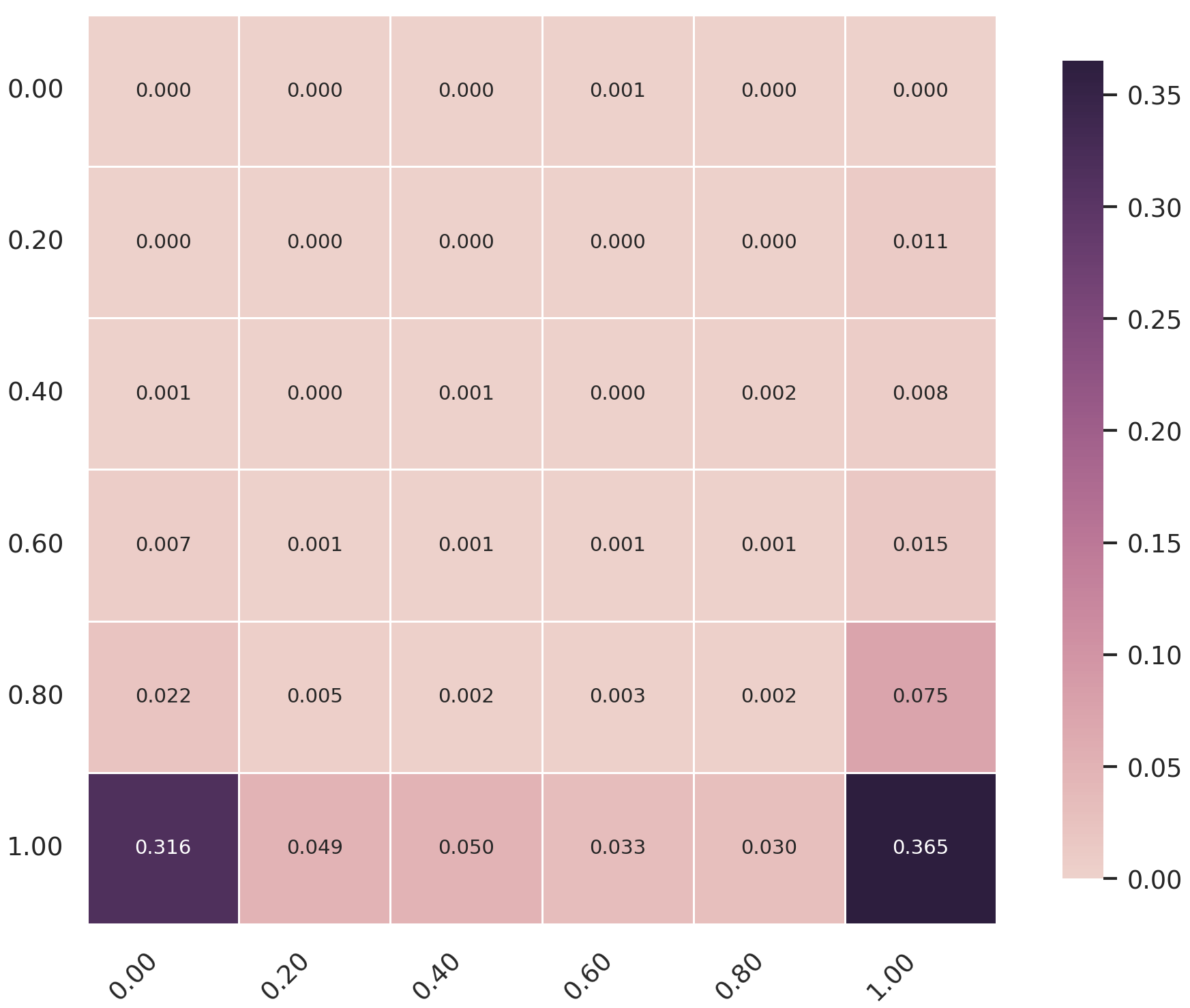} 
        \caption{FSTA vs. ITC - Hallucinations}
        \label{fig:plot4}
    \end{subfigure}
    
    \caption{\textbf{Joint probability density of Question-Aligned agreement metrics across consistency groups.} (a)-(b) show FSTC (y-axis) vs. FSTA (x-axis), and (c)-(d) show FSTA (y-axis) vs. ITC (x-axis) for the InstructBlip-7B model on the AMBER dataset.}
    \label{fig:joint_pmf}
\end{figure}

To demonstrate that hallucinations manifest as structured behavioral patterns within the consistency space, we visualize the joint probability density of QA metrics across complementary consistency groups in four plots (See Figure~\ref{fig:joint_pmf}). Each plot discretizes the joint space into bins where color intensity represents the percentage of samples in each bin. Crucially, the density patterns reveal that hallucinated and non-hallucinated samples occupy distinct regions in the consistency space. 

For example, the InstructBlip-7B model can hallucinate when it has low or high consistency on the ITC group, but if it has very high consistency on FSTA, it shows hallucination. The model even generates correct answers with low consistency in FSTA and hallucinated responses with high consistency in FSTA. However, when examining the joint pattern of FSTA and FSTC, we find the model is hallucinated when its consistency is high in FSTA and simultaneously low on FSTC. Thus, the joint pattern is more important than single group metrics, which prior works mostly rely on.

\subsection{Cross-Dataset Generalization}
To assess the generalizability of our method, we perform cross-dataset evaluations where the model is trained on one benchmark and tested on the other, as detailed in Tables~\ref{tab:cross_dataset}. The results demonstrate that our method exhibits robust transfer capabilities, maintaining competitive performance across both domain shifts.

\begin{table*}[!h]
\centering
\resizebox{\textwidth}{!}{%
\begin{tabular}{l ccc ccc ccc}
\toprule
\multirow{2}{*}{\textbf{Method}} 
& \multicolumn{3}{c}{\textbf{InstructBlip-7B}} 
& \multicolumn{3}{c}{\textbf{QWEN2.5-VL}} 
& \multicolumn{3}{c}{\textbf{InternVL-4B}} \\
\cmidrule(lr){2-4} \cmidrule(lr){5-7} \cmidrule(lr){8-10}
& \textbf{ACC} & \textbf{AUC-ROC} & \textbf{AUC-PR} 
& \textbf{ACC} & \textbf{AUC-ROC} & \textbf{AUC-PR} 
& \textbf{ACC} & \textbf{AUC-ROC} & \textbf{AUC-PR} \\
\midrule
\textbf{In-Distribution (AMBER $\to$ AMBER)} 
& \textbf{77.40} & \textbf{78.03} & \textbf{54.59} 
& \textbf{85.89} & \textbf{84.47} & \textbf{70.10} 
& \textbf{84.75} & \textbf{81.29} & \textbf{49.80} \\
\textbf{Cross-Dataset (PhD $\to$ AMBER)} 
& \textbf{73.50} & \textbf{73.44} & \textbf{51.43} 
& \textbf{76.50} & \textbf{80.54} & \textbf{59.78} 
& \textbf{84.50} & \textbf{77.73} & \textbf{42.20} \\
\midrule
\textbf{In-Distribution (PhD $\to$ PhD)} 
& \textbf{68.73} & \textbf{60.45} & \textbf{44.28} 
& \textbf{81.25} & \textbf{79.12} & \textbf{53.36} 
& \textbf{78.98} & \textbf{72.29} & \textbf{45.86} \\
\textbf{Cross-Dataset (AMBER $\to$ PhD)} 
& \textbf{63.80} & \textbf{60.33} & \textbf{40.59} 
& \textbf{72.50} & \textbf{67.83} & \textbf{38.04} 
& \textbf{76.10} & \textbf{69.52} & \textbf{40.04} \\
\bottomrule
\end{tabular}}
\caption{\textbf{Cross-dataset evaluation: Training on one and testing on another.}}
\label{tab:cross_dataset}
\end{table*}


\section{Limitations and Future work}

\textbf{Computational Cost.} Our framework operates entirely at inference time with minimal overhead: only a single lightweight classifier is trained once per LVLM, while hallucination detection relies on model outputs. Factual statement generation and transformation require 3 LLM inferences, followed by 30 LVLM inferences for the full configuration (Table~\ref{tab:sens_results}); the final classifier inference is negligible.

To furthur clarify the efficiency of UHP Detection, our minimal 10-inference setting matches the budget of our baselines. We therefore report an inference-budget comparison in Table~\ref{tab:budget_table}, constructed from Tables~\ref{tab:results-AMBER} and~\ref{tab:sens_results}. As shown in Table~\ref{tab:budget_table}, under a comparable 10–11 inference budget, our method consistently outperforms all baselines across both QWEN2.5-VL and InternVL-4B backbones. In particular, compared to QA-, BertScore-, Unigram-, and NLI-based verification methods, our approach achieves substantially higher accuracy and AUC scores without additional test-time computation, demonstrating a superior accuracy–efficiency trade-off.

\begin{table}[t]
\centering
\small
\setlength{\tabcolsep}{4pt}
\begin{tabular}{l|c|ccc|ccc}
\hline
Method & \#Inf. & \multicolumn{3}{c|}{QWEN2.5-VL} & \multicolumn{3}{c}{InternVL-4B} \\
 &  & ACC & AUC-ROC & AUC-PR & ACC & AUC-ROC & AUC-PR \\
\hline
QA & 10 & 62.23 & 64.48 & 33.08 & 66.40 & 54.90 & 30.34 \\
BertScore & 10 & 64.84 & 70.67 & 36.81 & 58.59 & 57.09 & 34.38 \\
Unigram & 10 & 65.62 & 65.86 & 33.96 & 61.45 & 56.63 & 33.50 \\
NLI & 10 & 77.86 & 63.61 & 34.04 & 64.06 & 64.14 & 36.40 \\
VL-Uncertainty & 11 & 66.25 & 61.83 & 24.25 & 68.75 & 69.02 & 35.38 \\
\textbf{Ours (I2-T2)} & 10 & \textbf{83.58} & \textbf{81.66} & \textbf{58.05} & \textbf{79.00} & \textbf{76.38} & \textbf{40.61} \\
\hline
\end{tabular}
\caption{Inference budget comparison on AMBER.}
\label{tab:budget_table}
\end{table}

\textbf{Applicability to Generative Tasks.} This work focuses on discriminative datasets (e.g., AMBER, PhD) where ground-truth hallucination labels are deterministically available via binary questions. Our framework is inherently extendable to generative tasks—if hallucination labels for generated text could be reliably obtained. Specifically, for any generated sentence, we could: (1) use original sentence as affirmative factual statement, (2) convert it to contradictory form, and (3) construct discriminative questions probing its core content. The critical barrier is the absence of accurate, automated hallucination labeling for generative outputs. Future work will prioritize developing an evaluator to generate these hallucination labels, enabling adaptation of UHP Detection to open-ended generation tasks. 


\section{Conclusion}
In this paper, we introduced UHP Detection, a fully black-box framework that redefines hallucination detection as identifying structured behavioral patterns within a multi-dimensional consistency space. By expanding the analysis across perturbation modalities and logical polarities, we demonstrated that hallucinations cannot be adequately characterized by single uncertainty metrics, but rather manifest as distinct configurations across fourteen within-group and between-group features. Extensive experiments on AMBER and PhD benchmarks using three LVLMs confirmed that our method consistently outperforms state-of-the-art baselines, achieving substantial gains in AUC-ROC and AUC-PR. Furthermore, ablation studies validated the complementary nature of visual and textual signals, and the effect of combining within- and between-group agreements features. Ultimately, the cross-dataset generalization and visualization of joint probability densities confirm that UHP Detection captures robust, transferable structures of model behavior for enhancing the reliability of Large Vision–Language Models by accurately identifying the \textbf{unique hallucination pattern}.


\clearpage


%
%
\bibliographystyle{tmlr}
\bibliography{main}


\clearpage
\appendix
\renewcommand{\thesection}{\Alph{section}}   
\renewcommand{\thesubsection}{\thesection.\arabic{subsection}}
\renewcommand{\thesubsubsection}{\thesubsection.\arabic{subsubsection}}

{
\setlength{\parindent}{0pt}
\begin{tcolorbox}[
    enhanced,
    colframe=ECCVBlue,       
    colback=SoftGray,        
    colbacktitle=ECCVBlue,   
    coltitle=white,          
    sharp corners=south,
    arc=5pt,
    drop fuzzy shadow,       
    fonttitle=\bfseries\Large,
    title={\faListOl\ Table of Contents},
    attach boxed title to top center={yshift=-2mm, yshifttext=-1mm},
    boxed title style={size=small, colframe=white, arc=3pt},
    left=5mm, right=5mm, top=8mm, bottom=6mm
]
    \renewcommand{\arraystretch}{1.5} 
    
    \begin{tabular}{@{}p{0.10\linewidth}@{}p{0.04\linewidth}@{}p{0.82\linewidth}@{}}
        \textbf{\color{ECCVBlue}\large A.} & 
        \color{ECCVBlue} & 
        {\textbf{\color{LinkColor}Prompt Construction}} \\
        & & \textcolor{gray!60}{\small Details on factual statement extraction and LVLM prompts.} \\[0.8em]
        
        \textbf{\color{ECCVBlue}\large B.} & 
        \color{ECCVBlue} & 
       {\textbf{\color{LinkColor}Image Transformations}} \\
        & & \textcolor{gray!60}{\small Visual examples of semantic-equivalent transformations.} \\[0.8em]
        
        \textbf{\color{ECCVBlue}\large C.} & 
        \color{ECCVBlue} & {\textbf{\color{LinkColor}Classifier Selection \& Hyperparameter Tuning}} \\
        & & \textcolor{gray!60}{\small Hyperparameter search spaces for linear and non-linear models.} \\[0.8em]
        
        \textbf{\color{ECCVBlue}\large D.} & 
        \color{ECCVBlue} & 
        {\textbf{\color{LinkColor}Ablation on Consistency Group Contributions}} \\
        & & \textcolor{gray!60}{\small Performance analysis of individual consistency features.} \\[0.8em]
        
        \textbf{\color{ECCVBlue}\large E.} & 
        \color{ECCVBlue} & 
        {\textbf{\color{LinkColor}Results on Hallucination Categories}} \\
        & & \textcolor{gray!60}{\small Hallucination category performance comparison.} \\
    \end{tabular}
\end{tcolorbox}
}

\clearpage


\section{Prompt Construction}
\label{sec:prompts}

\subsection{Factual Statement Construction}
For each discriminative sample, we provide a yes/no question and an image pair. We utilize the Phi-2 model to convert this binary question into affirmative and contradictory factual statements. The following prompts were used with Phi-2 to extract these statements.

\definecolor{TealDark}{RGB}{0, 100, 100}      
\definecolor{TealLight}{RGB}{230, 250, 250}   
\definecolor{OrangeDark}{RGB}{180, 70, 0}     
\definecolor{OrangeLight}{RGB}{255, 240, 230} 

\newtcolorbox{promptaffirmative}[1][]{
    enhanced,
    colback=white,
    colframe=white,           
    colbacktitle=TealDark,    
    coltitle=white,           
    borderline west={4pt}{0pt}{TealDark}, 
    fonttitle=\bfseries\small,
    title={\faCheckCircle\ #1}, 
    boxrule=0.5pt,
    arc=2pt,
    drop small lifted shadow,
    left=4mm, right=3mm, top=3mm, bottom=3mm,
    toptitle=2mm, bottomtitle=2mm,
    #1
}

\newtcolorbox{promptcontradictory}[1][]{
    enhanced,
    colback=white,
    colframe=white,           
    colbacktitle=OrangeDark,  
    coltitle=white,           
    borderline west={4pt}{0pt}{OrangeDark},
    fonttitle=\bfseries\small,
    title={\faCheckCircle\ #1}, 
    boxrule=0.5pt,
    arc=2pt,
    drop small lifted shadow,
    left=4mm, right=3mm, top=3mm, bottom=3mm,
    toptitle=2mm, bottomtitle=2mm,
    #1
}

\begin{promptaffirmative}{Affirmative Factual Statement Extraction Template}
    \ttfamily\scriptsize
    \setlength{\parskip}{0.6em}
    
    You are given a yes/no question, where the answer is \textbf{\textcolor{TealDark}{"yes"}}. 
    
    You should convert the question and the response to a single factual statement. As a result, the factual statement you generate should be a positive sentence. \\
    
    Your response must include \textbf{"only"} the factual statement, in the following format:
    
    \textbf{Factual Statement:} <factual statement>  \\
    
    \textbf{Example}  \\
    \textbf{Input:}  \\
    \textbf{Question:} Is the young boy in the image happy? \\
    \textbf{Output:}  \\
    Factual Statement: The young boy in the image is happy.  \\
    \\
    Now complete the task:  \\
    \textbf{Question:} \{question\}
\end{promptaffirmative}


\begin{promptcontradictory}{Contradictory Factual Statement Extraction Template}
    \ttfamily\scriptsize
    \setlength{\parskip}{0.6em}
    
    You are given a yes/no question, where the answer is \textbf{\textcolor{OrangeDark}{"no"}}.
    
    You should convert the question and the response to a single factual statement. As a result, the factual statement you generate should be a negative sentence.\\
    
    Your response must include \textbf{"only"} the factual statement, in the following format:
    
    \textbf{Factual Statement:} <factual statement>  \\
    
    \textbf{Example}  \\
    \textbf{Input:} \\
    \textbf{Question:} Is the young boy in the image happy \\
    \textbf{Output:} \\ 
    Factual Statement: The young boy in the image is not happy. \\
    \\
    Now complete the task: \\ 
    \textbf{Question:} \{question\}
\end{promptcontradictory}

\subsection{LVLM's Full Prompt Construction}
\label{app:full_prompt}

\begin{promptaffirmative}{Prompt construction (Affirmative factual statement)}
    \ttfamily\scriptsize
    \setlength{\parskip}{0.6em}
    \\ \\ Check the Image and following Claim then answer with True or False.
    
    \textbf{Claim:} There is a cat on the sofa.
    
    Is the Claim true or false?
\end{promptaffirmative}


\begin{promptcontradictory}{Prompt construction (Contradictory factual statement)}
    \ttfamily\scriptsize
    \setlength{\parskip}{0.6em}
    \\ \\ Check the Image and following Claim then answer with True or False.
    
    \textbf{Claim:} There is no cat on the sofa.
    
    Is the Claim true or false?
\end{promptcontradictory}

\clearpage

\section{Image Transformations}\label{sec:transforms}

\begin{figure}[!h]
    \centering
    \begin{subfigure}{0.3\textwidth}
        \centering
        \includegraphics[width=\linewidth,height=0.2\textheight,keepaspectratio]{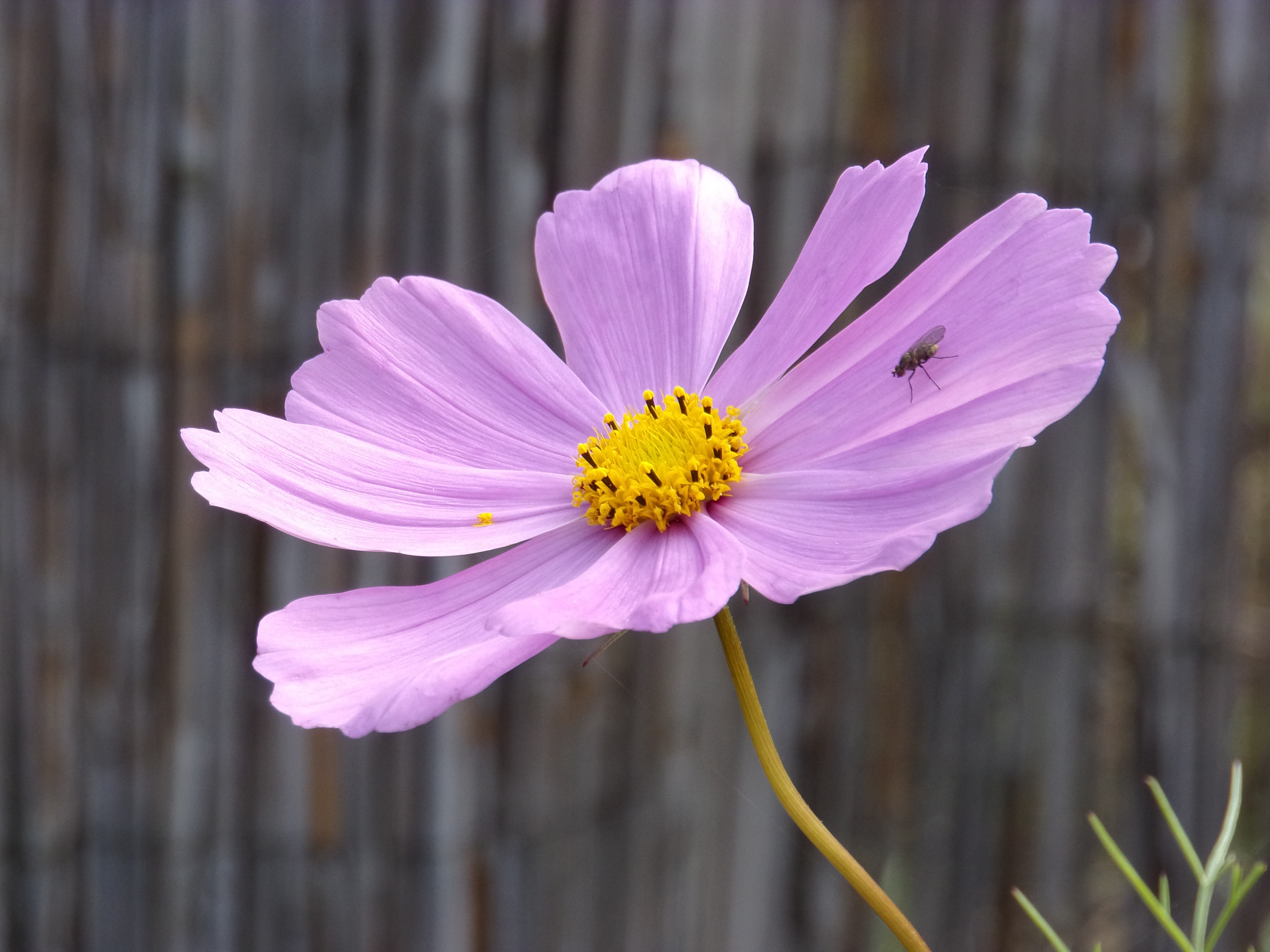}
        \caption{\textbf{Original Image}}
        \label{fig:orig}
    \end{subfigure}
    \hfill
    \begin{subfigure}{0.3\textwidth}
        \centering
        \includegraphics[width=\linewidth,height=0.20\textheight,keepaspectratio]{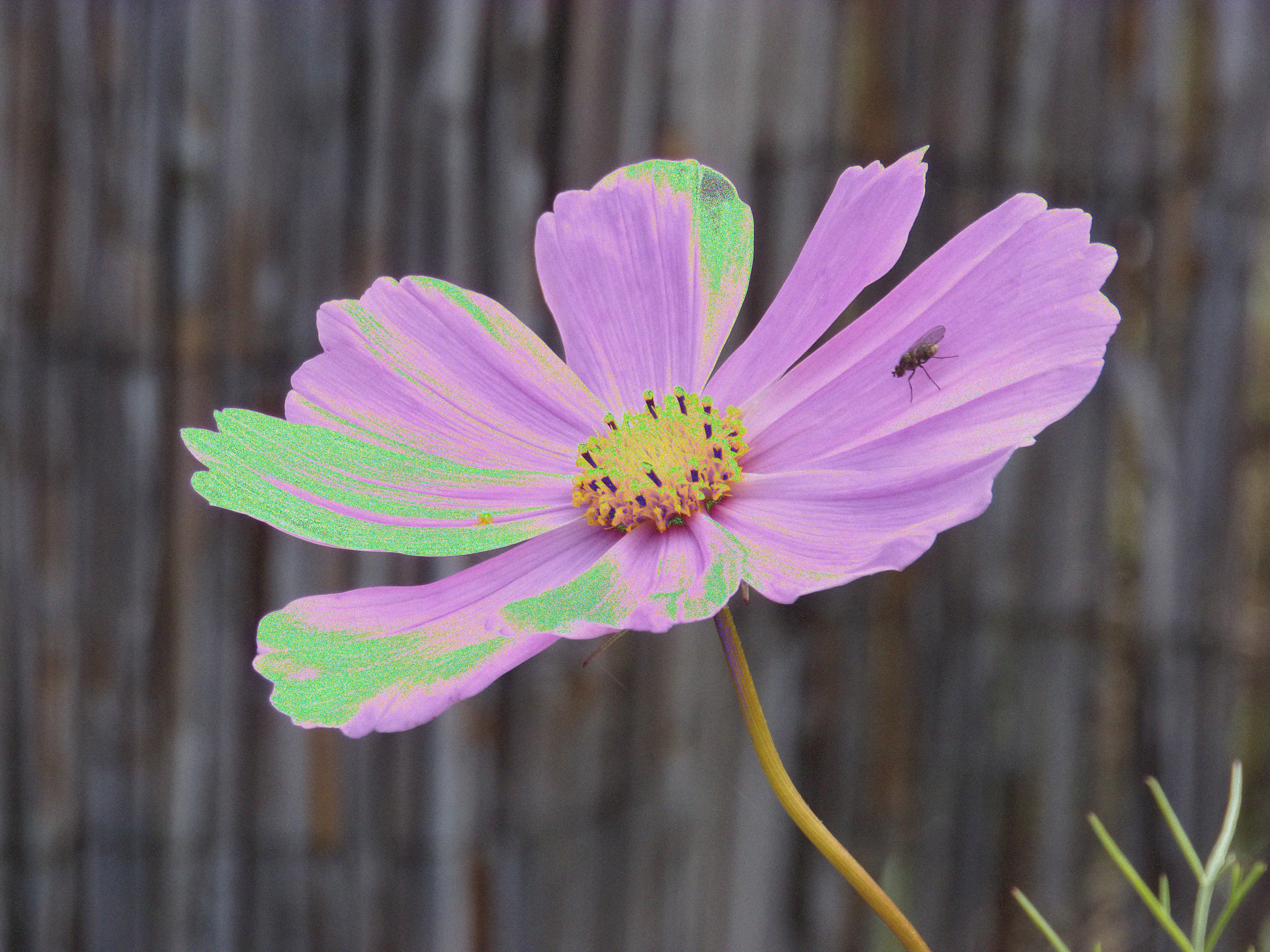}
        \caption{Gaussian noise (1.0)}
        \label{fig:trans1}
    \end{subfigure}
    \hfill
    \begin{subfigure}{0.3\textwidth}
        \centering
        \includegraphics[width=\linewidth,height=0.20\textheight,keepaspectratio]{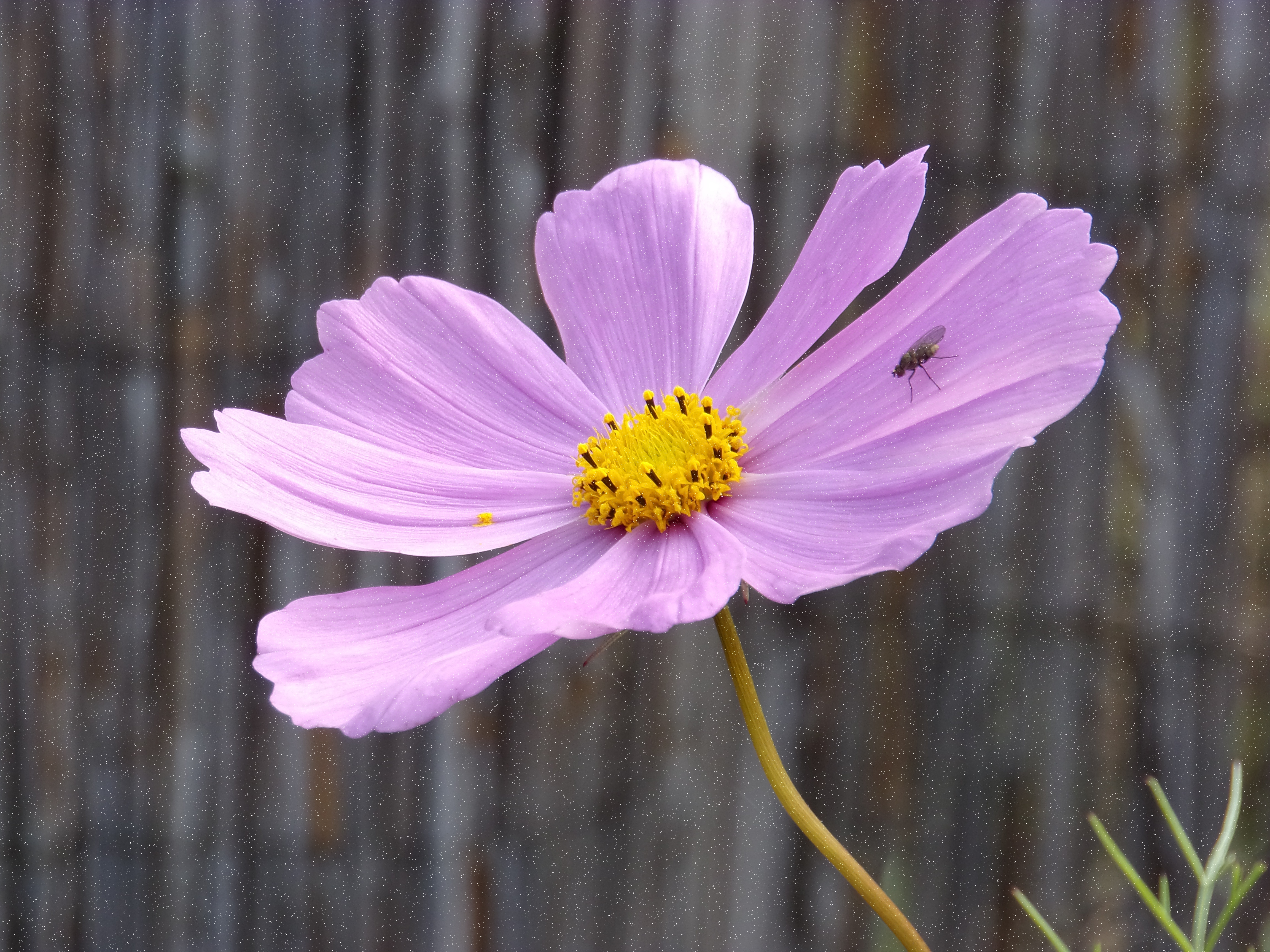}
        \caption{Impulse noise (1.0)}
        \label{fig:trans2}
    \end{subfigure}
    
    \vspace{1em}
    \begin{subfigure}{0.3\textwidth}
        \centering
        \includegraphics[width=\linewidth,height=0.20\textheight,keepaspectratio]{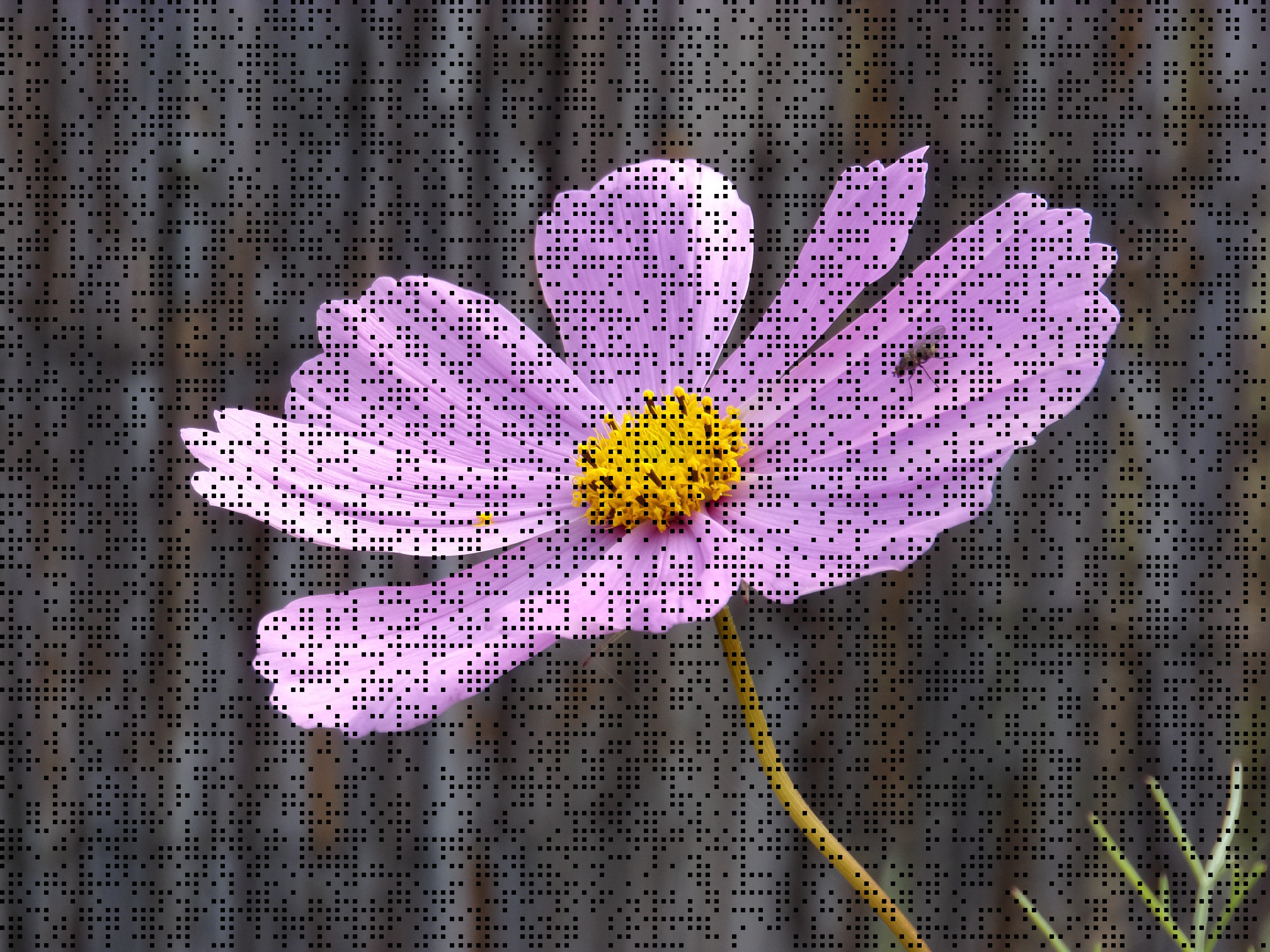}
        \caption{Grid mask (0.9995)}
        \label{fig:trans3}
    \end{subfigure}
    \hfill
    \begin{subfigure}{0.3\textwidth}
        \centering
        \includegraphics[width=\linewidth,height=0.20\textheight,keepaspectratio]{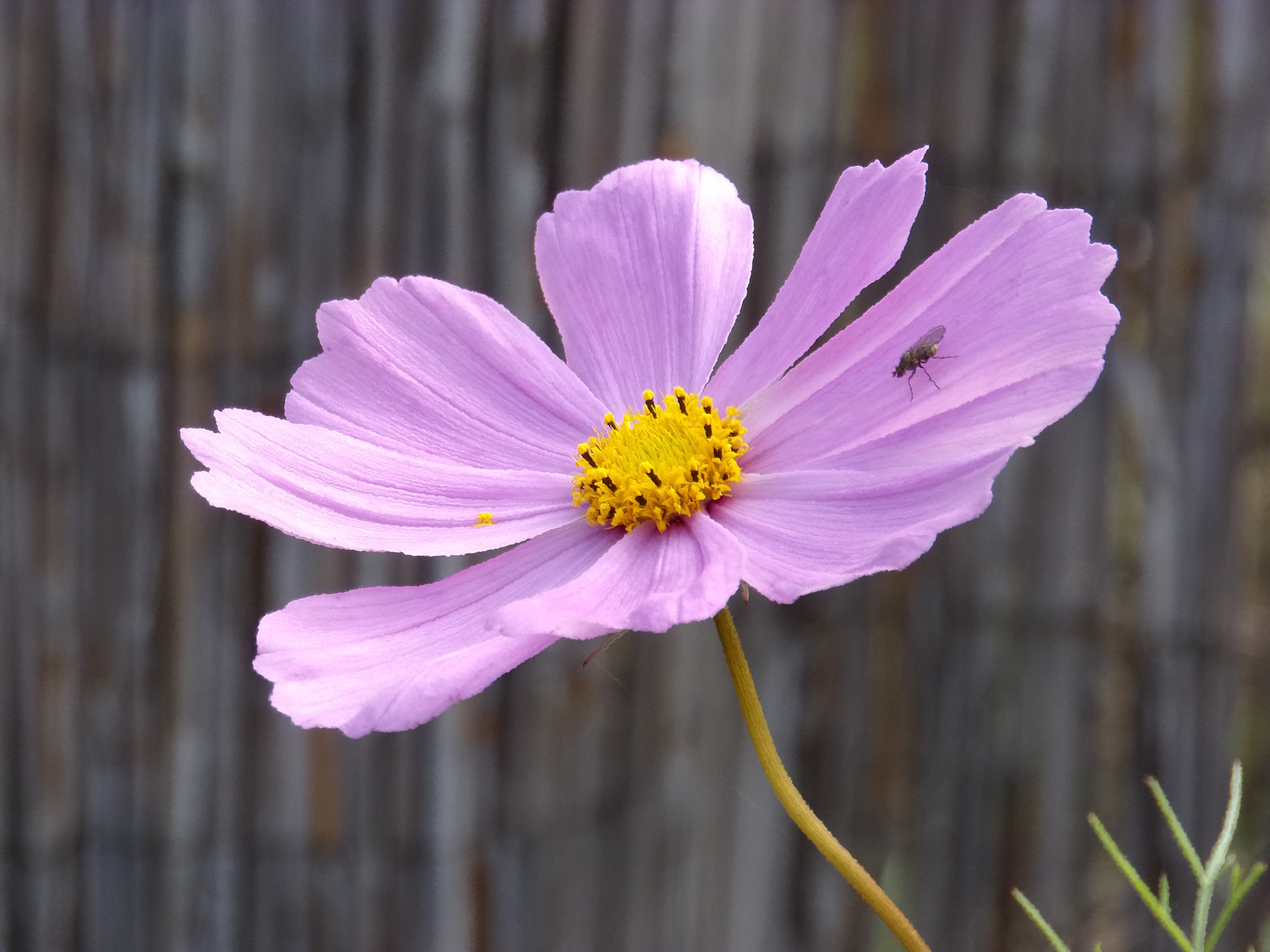}
        \caption{Elastic transform (1.0)}
        \label{fig:trans4}
    \end{subfigure}
    \hfill
    \begin{subfigure}{0.3\textwidth}
        \centering
        \includegraphics[width=\linewidth,height=0.20\textheight,keepaspectratio]{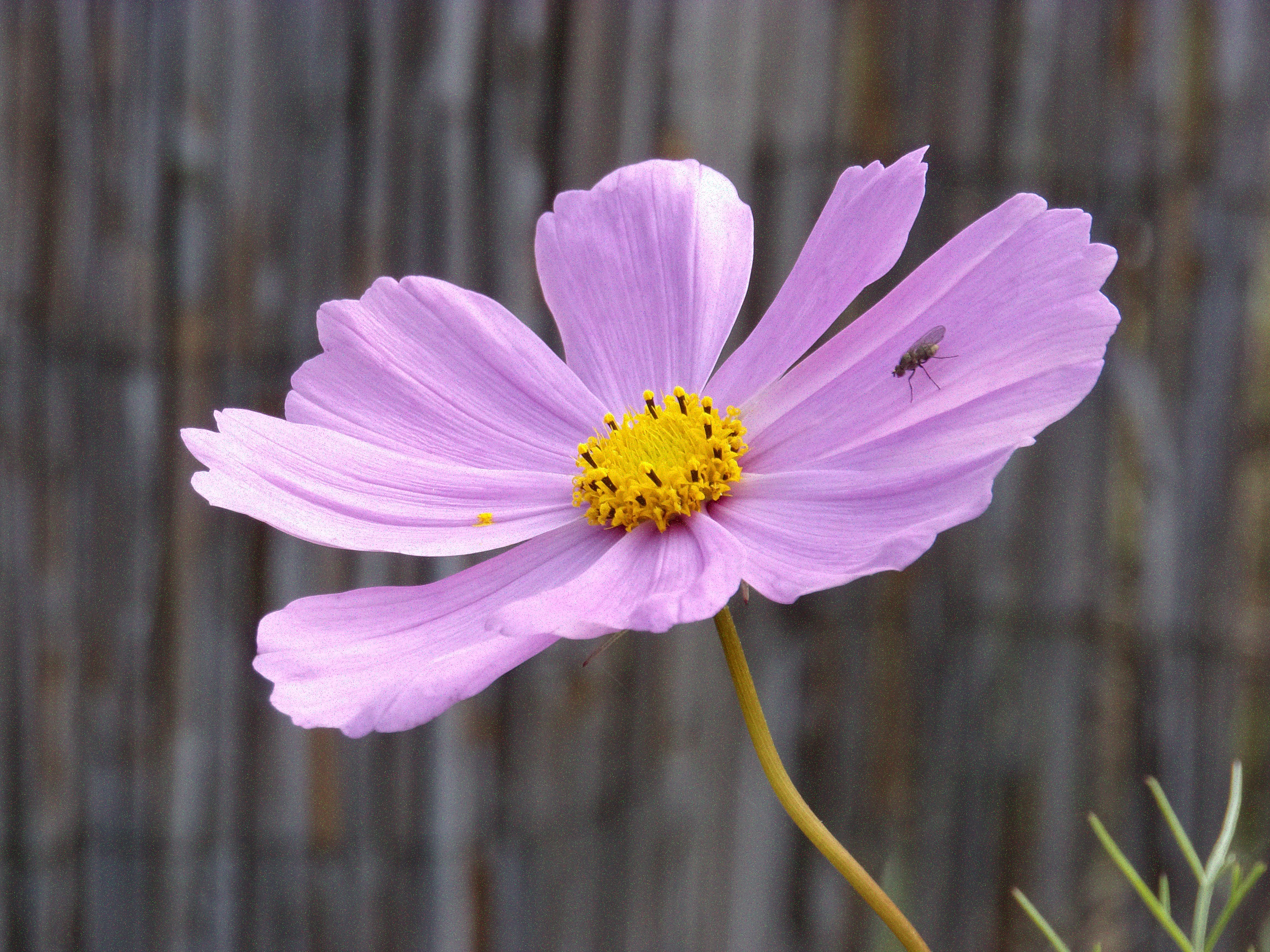}
        \caption{Salt \& pepper (0.9998)}
        \label{fig:trans5}
    \end{subfigure}
    
    \vspace{1em}
    \begin{subfigure}{0.3\textwidth}
        \centering
        \includegraphics[width=\linewidth,height=0.20\textheight,keepaspectratio]{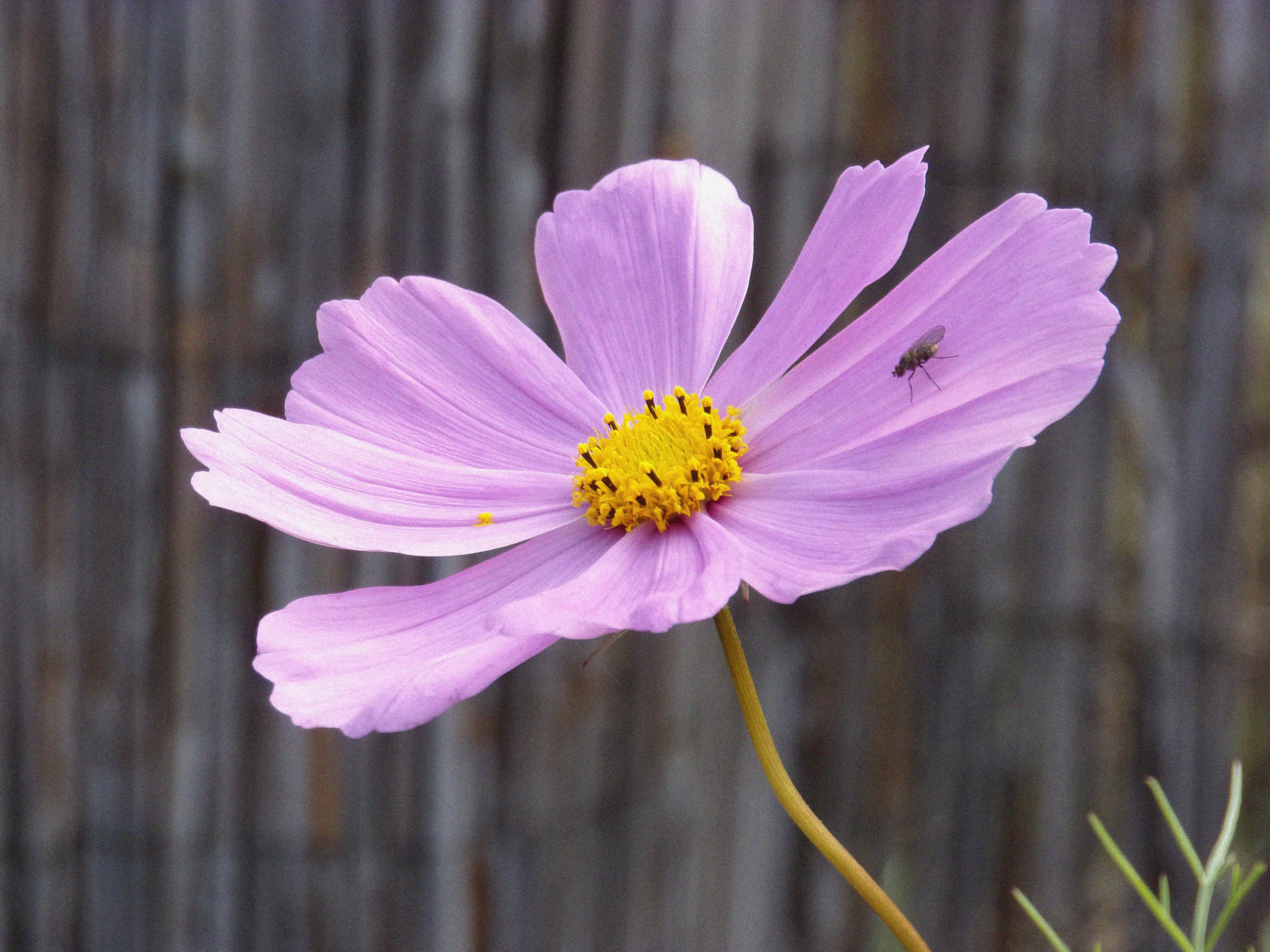}
        \caption{Augmentation (0.9987)}
        \label{fig:trans6}
    \end{subfigure}
    \hfill
    \begin{subfigure}{0.3\textwidth}
        \centering
        \includegraphics[width=\linewidth,height=0.20\textheight,keepaspectratio]{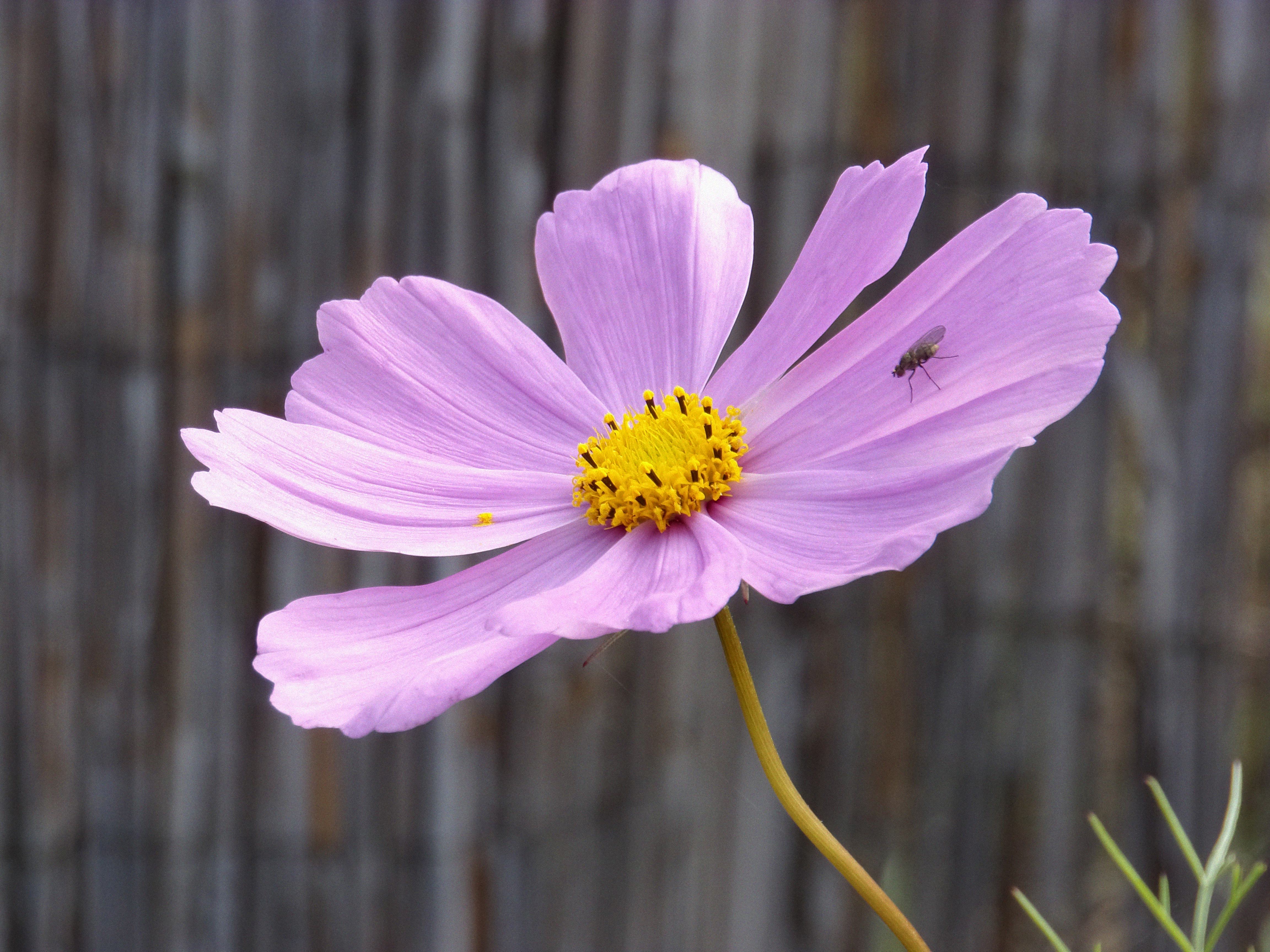}
        \caption{Poisson noise (1.0)}
        \label{fig:trans7}
    \end{subfigure}
    \hfill
    \begin{subfigure}{0.3\textwidth}
        \centering
        \includegraphics[width=\linewidth,height=0.20\textheight,keepaspectratio]{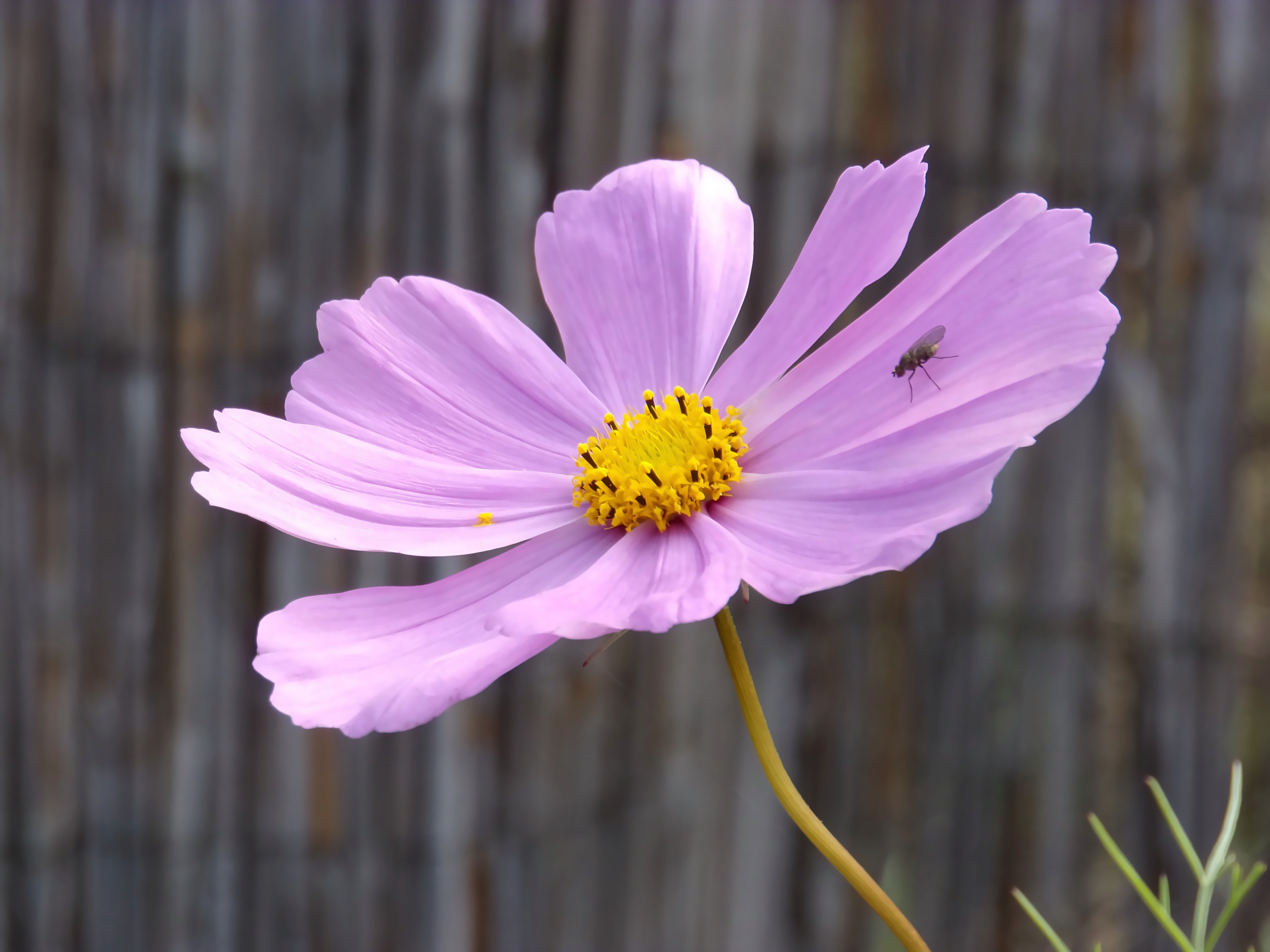}
        \caption{Non-local means (1.0)}
        \label{fig:trans8}
    \end{subfigure}
    
    \vspace{1em}
    \begin{subfigure}{0.3\textwidth}
        \centering
        \includegraphics[width=\linewidth,height=0.20\textheight,keepaspectratio]{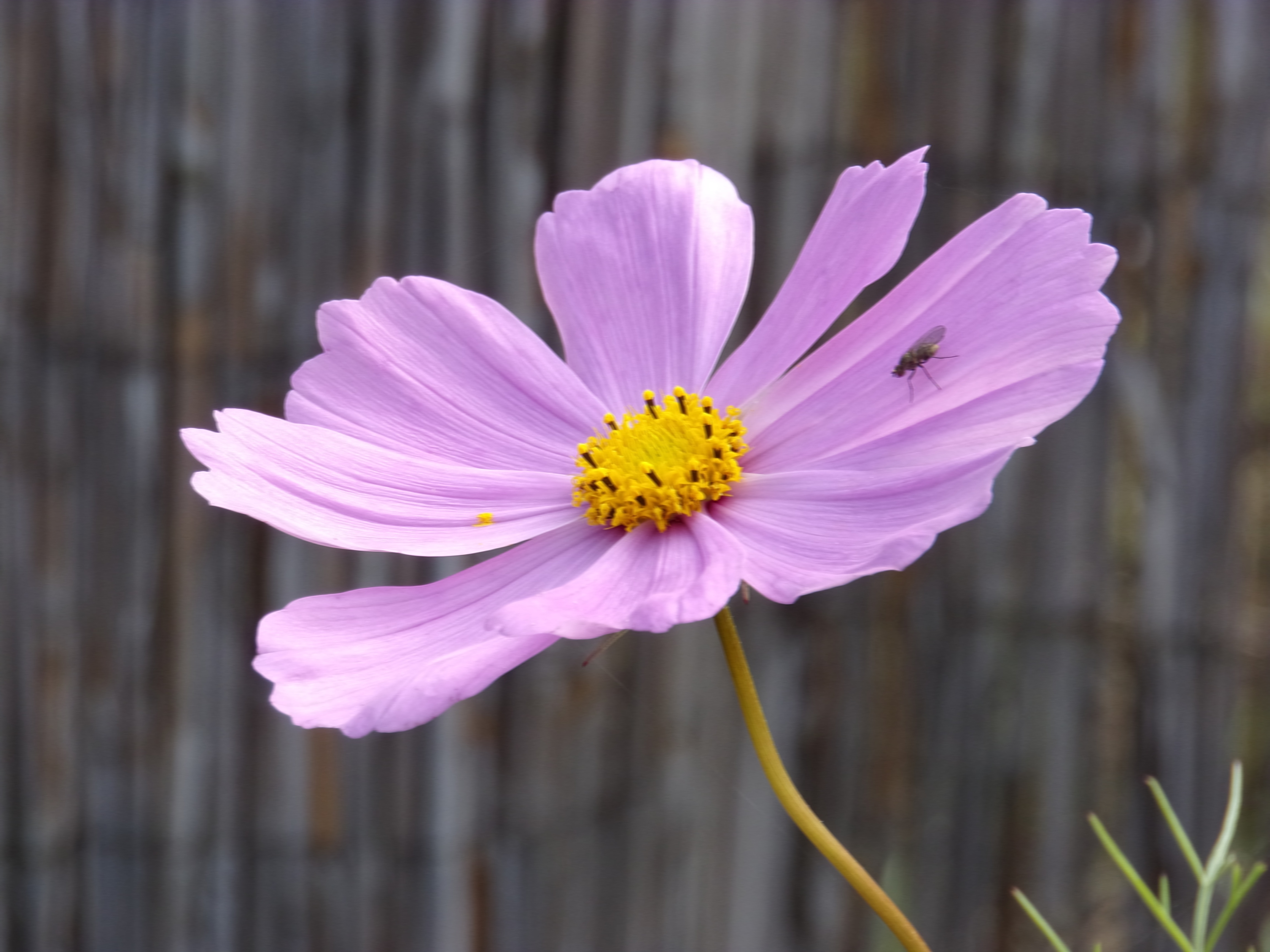}
        \caption{Motion blur (0.9899)}
        \label{fig:trans9}
    \end{subfigure}
    \hspace{1em}
    \begin{subfigure}{0.3\textwidth}
        \centering
        \includegraphics[width=\linewidth,height=0.20\textheight,keepaspectratio]{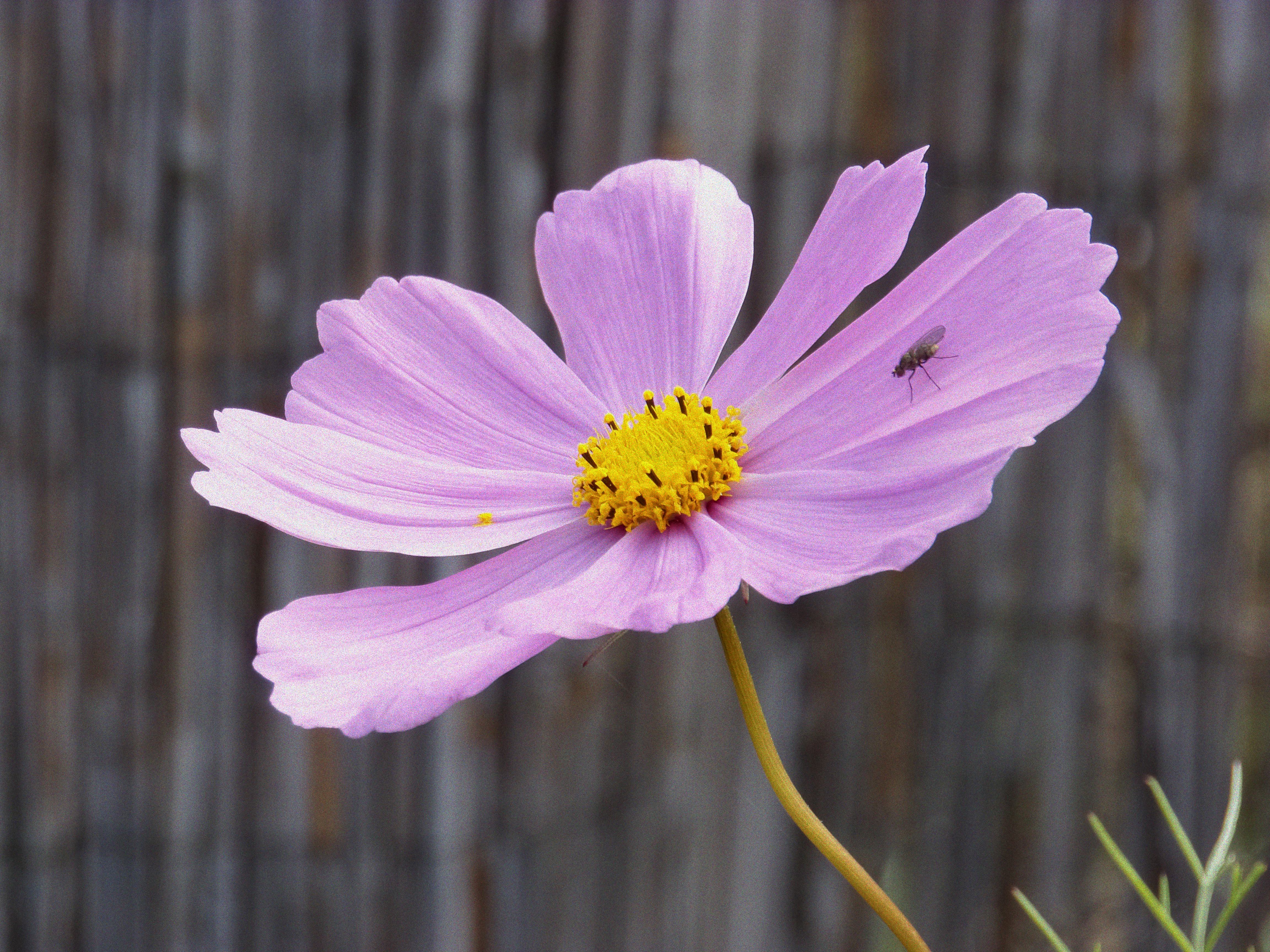}
        \caption{Speckle noise (0.9999)}
        \label{fig:trans10}
    \end{subfigure}

    \caption{\textbf{Semantically equivalent image transformations.} The original image (top left) is followed by 10 transformations applied to the same input, with scene composition similarity scores reported in parentheses. High similarity scores demonstrate that these transformations preserve the original factual content.}
    \label{fig:transformations}
\end{figure}

\clearpage

\section{Classifier Selection and Hyperparameter Tuning}\label{sec:classifiers}

Below, we summarize the selected classifiers and their corresponding search spaces.

\paragraph{Logistic Regression.}
A linear classifier with $\ell_2$ regularization. We include feature standardization and class balancing.
\begin{itemize}
    \item Regularization strength: $C \in \{0.01, 0.1, 1, 10\}$
    \item Penalty: $\ell_2$
    \item Max iterations: 2000
    \item Class weight: balanced
\end{itemize}

\paragraph{Ridge Classifier.}
A linear classifier using $\ell_2$ penalty in a least-squares formulation.
\begin{itemize}
    \item Regularization parameter: $\alpha \in \{0.1, 1, 10\}$
\end{itemize}

\paragraph{Support Vector Machine (RBF Kernel).}
A non-linear kernel method with probabilistic outputs enabled.
\begin{itemize}
    \item Regularization parameter: $C \in \{0.1, 1, 10\}$
    \item Kernel coefficient: $\gamma \in \{\text{scale}, \text{auto}\}$
    \item Class weight: balanced
\end{itemize}

\paragraph{Decision Tree.}
A single CART decision tree with cost-complexity pruning.
\begin{itemize}
    \item Max depth: $\{ \text{None}, 3, 5, 10\}$
    \item Min samples split: $\{2, 5, 10\}$
    \item Pruning parameter: $ccp\_alpha \in \{0.0, 0.01, 0.1\}$
\end{itemize}

\paragraph{Random Forest.}
An ensemble of decision trees trained with bootstrap aggregation.
\begin{itemize}
    \item Number of trees: $\{100, 200\}$
    \item Max depth: $\{\text{None}, 5, 10\}$
    \item Min samples split: $\{2, 5, 10\}$
    \item Max features: $\{\text{sqrt}, \text{log2}\}$
    \item Class weight: balanced
\end{itemize}

\paragraph{Gradient Boosting.}
Stage-wise additive tree ensemble optimized via gradient descent.
\begin{itemize}
    \item Number of estimators: $\{100, 200\}$
    \item Learning rate: $\{0.05, 0.1, 0.2\}$
    \item Max depth: $\{3, 5\}$
    \item Subsample: $\{0.8, 1.0\}$
\end{itemize}

\paragraph{AdaBoost.}
Boosting ensemble with adaptive sample reweighting.
\begin{itemize}
    \item Number of estimators: $\{50, 100, 200\}$
    \item Learning rate: $\{0.01, 0.1, 1\}$
\end{itemize}

\paragraph{XGBoost (Tree Booster).}
Gradient-boosted decision trees with second-order optimization.
\begin{itemize}
    \item Number of estimators: $\{100, 200\}$
    \item Learning rate: $\{0.05, 0.1, 0.2\}$
    \item Max depth: $\{3, 5, 7\}$
    \item Subsample: $\{0.8, 1.0\}$
    \item Evaluation metric: log-loss
\end{itemize}

\paragraph{XGBoost (Linear Booster).}
Linear booster variant for comparison with classical linear models.
\begin{itemize}
    \item Learning rate: $\{0.01, 0.05, 0.1\}$
    \item $\ell_1$ regularization: $\text{reg\_alpha} \in \{0, 0.1, 1\}$
    \item $\ell_2$ regularization: $\text{reg\_lambda} \in \{0, 0.1, 1\}$
\end{itemize}

\clearpage

\section{Ablation on Consistency Group Contributions}

We conducted ablation studies to evaluate the contribution of individual components in our hallucination detection framework. Specifically, we systematically removed the following elements across both AMBER and PhD datasets:
(1) Individual consistency groups (ITA, ITC, FSTA, FSTC),
(2) Modalities (Image-only, Text-only), and
(3) Logical polarity components (Affirmative-only, Contradictory-only).
For each ablation variant, we measured performance using accuracy (ACC), AUC-ROC, and AUC-PR metrics. The full UHP Detection model (which retains all components) served as the baseline for comparison. Results are presented in Tables~\ref{tab:ablation_amber_final} and~\ref{tab:ablation_phd_final}.
\label{sec:ablation}

\definecolor{tablegray}{RGB}{240, 240, 240}

\begin{table*}[!h]
\centering
\setlength{\tabcolsep}{3.5pt}
\renewcommand{\arraystretch}{1.15}
\resizebox{\textwidth}{!}{
\begin{tabular}{l ccc ccc ccc}
\toprule
\multirow{2}{*}{\textbf{Method}}
& \multicolumn{3}{c}{\textbf{InstructBlip-7B}}
& \multicolumn{3}{c}{\textbf{Qwen2.5-VL}}
& \multicolumn{3}{c}{\textbf{InternVL-4B}} \\
\cmidrule(lr){2-4} \cmidrule(lr){5-7} \cmidrule(lr){8-10}
& ACC & AUC-ROC & AUC-PR
& ACC & AUC-ROC & AUC-PR
& ACC & AUC-ROC & AUC-PR \\
\midrule
w/o ITA
& 74.75\down{2.65} & 73.23\down{4.80} & 49.51\down{5.08}
& 84.61\down{1.28} & 81.17\down{3.30} & 66.89\down{3.21}
& 79.50\down{5.25} & 79.71\down{1.71} & 48.49\down{1.31} \\
\rowcolor{tablegray}w/o ITC
& 73.75\down{3.65} & 69.65\down{8.38} & 48.10\down{6.49}
& 83.33\down{2.56} & 80.95\down{3.52} & 65.76\down{4.34}
& 80.25\down{4.50} & 77.01\down{4.41} & 45.92\down{3.88} \\
w/o FSTA
& 74.50\down{2.90} & 72.17\down{5.86} & 49.11\down{5.48}
& 83.07\down{2.82} & 79.18\down{5.29} & 57.98\down{12.12}
& 83.75\down{1.00} & 78.73\down{2.56} & 45.89\down{3.91} \\
\rowcolor{tablegray}w/o FSTC
& 75.25\down{2.15} & 73.28\down{4.75} & 47.69\down{6.90}
& 82.05\down{3.84} & 83.01\down{1.46} & 65.19\down{4.91}
& 84.00\down{0.75} & 80.76\down{0.53} & 47.03\down{2.77} \\
Image-only
& 69.25\down{8.15} & 72.01\down{6.02} & 48.00\down{6.59}
& 81.02\down{4.87} & 80.09\down{4.38} & 60.75\down{9.35}
& 76.75\down{8.00} & 78.49\down{2.80} & 43.11\down{6.69} \\
\rowcolor{tablegray}Text-only
& 67.50\down{9.90} & 68.25\down{9.78} & 46.25\down{8.34}
& 84.35\down{1.54} & 82.03\down{2.44} & 63.17\down{6.93}
& 74.00\down{10.75} & 76.14\down{5.15} & 40.01\down{9.79} \\
Affirmative only
& 72.50\down{4.90} & 71.69\down{6.34} & 48.45\down{6.14}
& 76.66\down{9.23} & 73.45\down{11.02} & 56.46\down{13.64}
& 77.75\down{7.00} & 78.48\down{2.81} & 41.61\down{8.19} \\
\rowcolor{tablegray}Contradictory only
& 72.50\down{4.90} & 70.03\down{8.00} & 45.25\down{9.34}
& 83.07\down{2.82} & 78.98\down{5.49} & 63.83\down{6.27}
& 76.25\down{8.50} & 72.94\down{8.35} & 40.64\down{9.16} \\
\midrule
\rowcolor{lightblue}\textbf{UHP Detection (Full)}
& \textbf{77.40}
& \textbf{78.03}
& \textbf{54.59}
& \textbf{85.89}
& \textbf{84.47}
& \textbf{70.10}
& \textbf{84.75}
& \textbf{81.29}
& \textbf{49.80} \\
\bottomrule
\end{tabular}}
\caption{\textbf{Ablation study on the AMBER dataset.} Performance drops ($\downarrow$) are reported relative to the full configuration.}
\label{tab:ablation_amber_final}
\end{table*}

\begin{table*}[!h]
\centering
\setlength{\tabcolsep}{3.5pt}
\renewcommand{\arraystretch}{1.15}
\resizebox{\textwidth}{!}{%
\begin{tabular}{l ccc ccc ccc}
\toprule
\multirow{2}{*}{\textbf{Method}} 
& \multicolumn{3}{c}{\textbf{InstructBlip-7B}} 
& \multicolumn{3}{c}{\textbf{QWEN2.5-VL}} 
& \multicolumn{3}{c}{\textbf{InternVL-4B}} \\
\cmidrule(lr){2-4} \cmidrule(lr){5-7} \cmidrule(lr){8-10}
& ACC & AUC-ROC & AUC-PR 
& ACC & AUC-ROC & AUC-PR 
& ACC & AUC-ROC & AUC-PR \\
\midrule
w/o ITA
& 60.50\down{8.23} & 57.85\down{2.60} & 39.23\down{5.05}
& 80.25\down{1.00} & 74.71\down{4.41} & 48.92\down{4.44}
& 77.75\down{1.23} & 70.36\down{1.93} & 43.47\down{2.39} \\
\rowcolor{tablegray}w/o ITC
& 60.75\down{7.98} & 58.32\down{2.13} & 39.19\down{5.09}
& 80.75\down{0.50} & 77.10\down{2.02} & 47.81\down{5.55}
& 77.00\down{1.98} & 68.94\down{3.35} & 41.61\down{4.25} \\
w/o FSTA
& 60.00\down{8.73} & 57.08\down{3.37} & 36.15\down{8.13}
& 79.75\down{1.50} & 77.71\down{1.41} & 49.16\down{4.20}
& 68.75\down{10.23} & 63.30\down{8.99} & 40.24\down{5.62} \\
\rowcolor{tablegray}w/o FSTC
& 57.25\down{11.48} & 57.30\down{3.15} & 37.85\down{6.43}
& 79.25\down{2.00} & 76.68\down{2.44} & 49.38\down{3.98}
& 65.75\down{13.23} & 66.84\down{5.45} & 41.08\down{4.78} \\
Image-only
& 54.50\down{14.23} & 50.47\down{9.98} & 35.68\down{8.60}
& 73.50\down{7.75} & 76.48\down{2.64} & 44.46\down{8.90}
& 67.50\down{11.48} & 63.91\down{8.38} & 39.69\down{6.17} \\
\rowcolor{tablegray}Text-only
& 56.25\down{12.48} & 55.69\down{4.76} & 36.35\down{7.93}
& 74.75\down{6.50} & 76.88\down{2.24} & 45.01\down{8.35}
& 66.75\down{12.23} & 64.85\down{7.44} & 38.19\down{7.67} \\
Affirmative only
& 52.25\down{16.48} & 57.09\down{3.36} & 37.56\down{6.72}
& 75.25\down{6.00} & 72.45\down{6.67} & 42.20\down{11.16}
& 66.25\down{12.73} & 63.97\down{8.32} & 38.30\down{7.56} \\
\rowcolor{tablegray}Contradictory only
& 58.75\down{9.98} & 54.78\down{5.67} & 35.87\down{8.41}
& 74.75\down{6.50} & 75.40\down{3.72} & 43.28\down{10.08}
& 64.75\down{14.23} & 62.09\down{10.20} & 34.08\down{11.78} \\
\midrule
\rowcolor{lightblue}\textbf{UHP Detection (Full)} 
& \textbf{68.73}
& \textbf{60.45}
& \textbf{44.28}
& \textbf{81.25}
& \textbf{79.12}
& \textbf{53.36}
& \textbf{78.98}
& \textbf{72.29}
& \textbf{45.86} \\
\bottomrule
\end{tabular}}
\caption{\textbf{Ablation study on the PhD dataset.} Performance drops ($\downarrow$) are reported relative to the full configuration.}
\label{tab:ablation_phd_final}
\end{table*}

\clearpage

\section{Results on Hallucination Categories}\label{sec:categories}

\begin{figure}[!h]
    \centering
    \includegraphics[height=0.28\textheight, width=\textwidth, keepaspectratio]{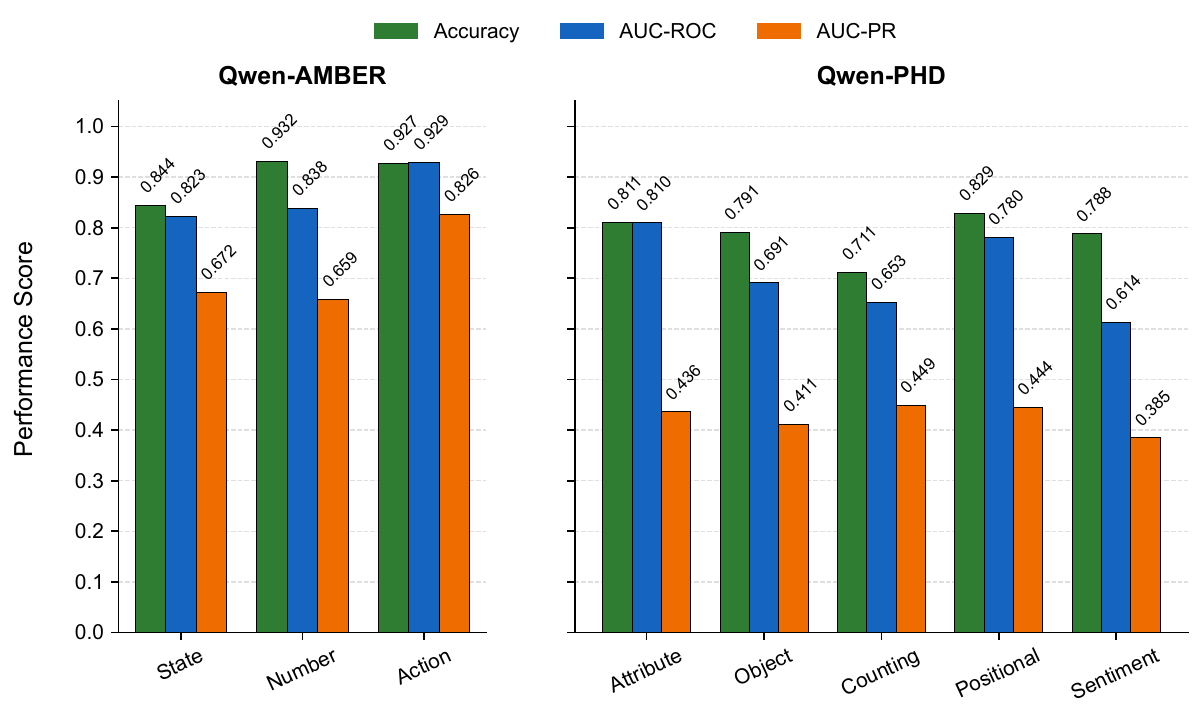}
    \vspace{0.1em}
    
    \includegraphics[height=0.28\textheight, width=\textwidth, keepaspectratio]{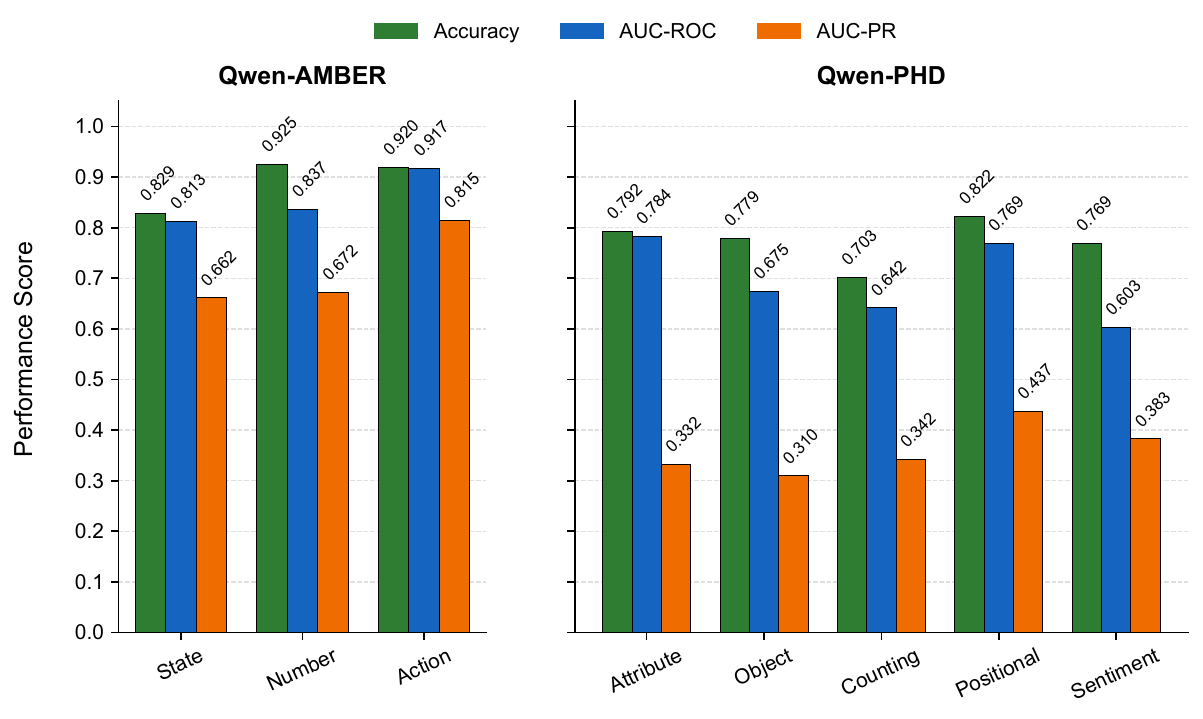}
    \vspace{0.1em}
    
    \includegraphics[height=0.28\textheight, width=\textwidth, keepaspectratio]{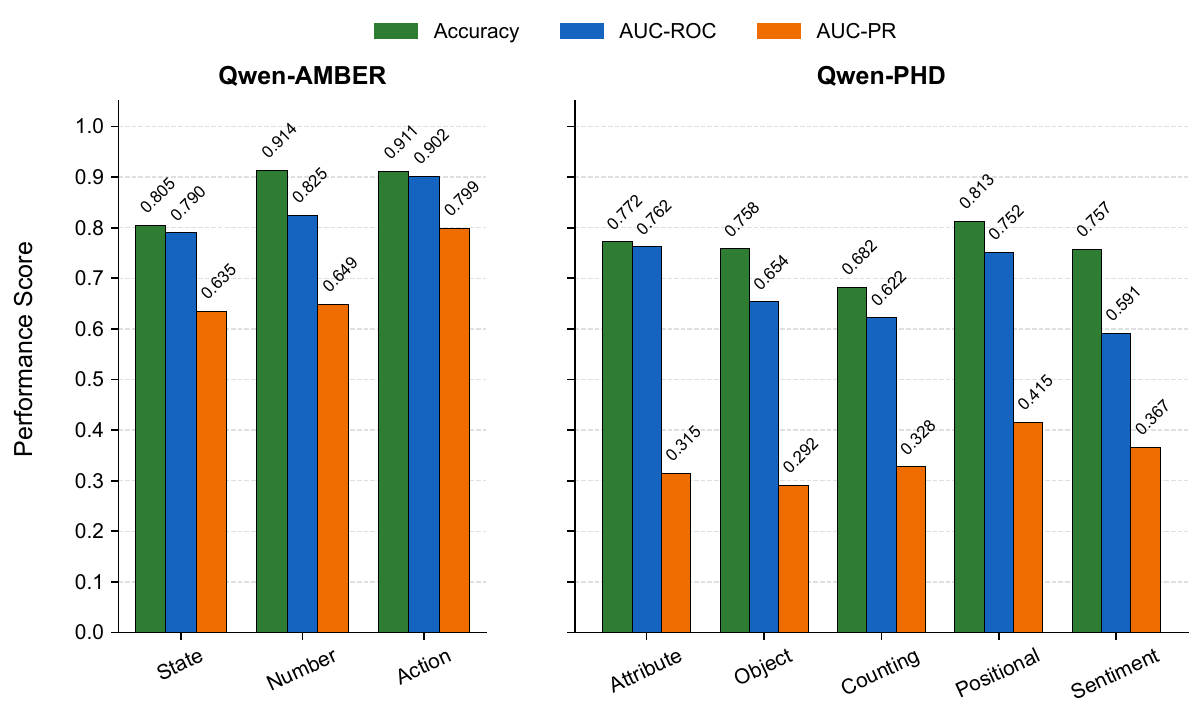}
    
    \caption{
    Hallucination category performance comparison across detection methods.
    Top to bottom: Our Method, NLI, and Unigram.
    }
    \label{fig:category_results}
\end{figure}

\end{document}